%% file: 4358.tex
\documentclass{article}
\input{math_commands.tex}

\usepackage[preprint]{neurips_2025}

\usepackage[utf8]{inputenc} 
\usepackage[T1]{fontenc}    
\usepackage[dvipsnames]{xcolor}
\usepackage{hyperref}       
\hypersetup{
    colorlinks=true,
	citecolor=MidnightBlue,
	linkcolor=OrangeRed,
	urlcolor=OrangeRed
}
\usepackage{url}            
\usepackage{booktabs}       
\usepackage{amsfonts}       
\usepackage{nicefrac}       
\usepackage{microtype}      
\usepackage{xcolor}         

\usepackage{todonotes}
\usepackage{amsthm}
\usepackage{amsmath}
\usepackage[capitalise,nameinlink]{cleveref}
\usepackage{amssymb}
\usepackage{makecell}
\usepackage{multirow}
\usepackage{footnote}
\usepackage{subcaption}
\usepackage{graphicx}
\usepackage{wrapfig}
\usepackage{bbding}
\usepackage{todonotes}
\usepackage{algorithmic}
\usepackage{algorithm}
\usepackage{bigstrut}
\usepackage{tabularx}
\usepackage{makecell}
\usepackage{titletoc}

\title{Quantize-then-Rectify: Efficient VQ-VAE Training}

\newcommand*\samethanks[1][\value{footnote}]{\footnotemark[#1]}
\author{Borui Zhang\thanks{Equal contribution.}\ , Qihang Rao\samethanks\ , Wenzhao Zheng\ , Jie Zhou\ , Jiwen Lu\thanks{Corresponding author.}\\
{Department of Automation, Tsinghua University, China}\\
}

\begin{document}

\maketitle

\input{chapters/0_abstract.tex}
\input{chapters/1_introduction.tex}
\input{chapters/2_related_work.tex}
\input{chapters/3_method.tex}
\input{chapters/4_experiment.tex}
\input{chapters/5_conclusion.tex}

{
    \small
    \bibliographystyle{plainnat}
    \bibliography{ref}
}
\input{chapters/appendix.tex}

\end{document}

%% file: math_commands.tex
\usepackage{amsmath,amsfonts,bm}

\def\eqref#1{equation~\ref{#1}}

\def\1{\bm{1}}

\def\vc{{\bm{c}}}

\def\vx{{\bm{x}}}

\def\vz{{\bm{z}}}

\def\mZ{{\bm{Z}}}

\DeclareMathAlphabet{\mathsfit}{\encodingdefault}{\sfdefault}{m}{sl}
\SetMathAlphabet{\mathsfit}{bold}{\encodingdefault}{\sfdefault}{bx}{n}

\def\sI{{\mathbb{I}}}

\def\sZ{{\mathbb{Z}}}

\DeclareMathOperator*{\argmin}{arg\,min}

%% file: chapters/0_abstract.tex
\begin{abstract}
    Visual tokenizers are pivotal in multimodal large models, acting as bridges between continuous inputs and discrete token. 
    Nevertheless, training high-compression-rate VQ-VAEs remains computationally demanding, often necessitating thousands of GPU hours. 
    This work demonstrates that a pre-trained VAE can be efficiently transformed into a VQ-VAE by controlling quantization noise within the VAE's tolerance threshold. 
    We present \textbf{Quantize-then-Rectify (ReVQ)}, a framework leveraging pre-trained VAEs to enable rapid VQ-VAE training with minimal computational overhead. 
    By integrating \textbf{channel multi-group quantization} to enlarge codebook capacity and a \textbf{post rectifier} to mitigate quantization errors, ReVQ compresses ImageNet images into at most $512$ tokens while sustaining competitive reconstruction quality (rFID = $1.06$). 
    Significantly, ReVQ reduces training costs by over two orders of magnitude relative to state-of-the-art approaches: 
    ReVQ finishes full training on a single NVIDIA 4090 in approximately 22 hours, whereas comparable methods require 4.5 days on a 32 A100 GPUs. 
    Experimental results show that ReVQ achieves superior efficiency-reconstruction trade-offs.
    \footnote{Code: \url{https://github.com/Neur-IO/ReVQ}}
\end{abstract}

%% file: chapters/1_introduction.tex
\vspace{-3mm}
\section{Introduction}

\begin{figure}[htbp]
    \centering
    \includegraphics[width=0.95\linewidth]{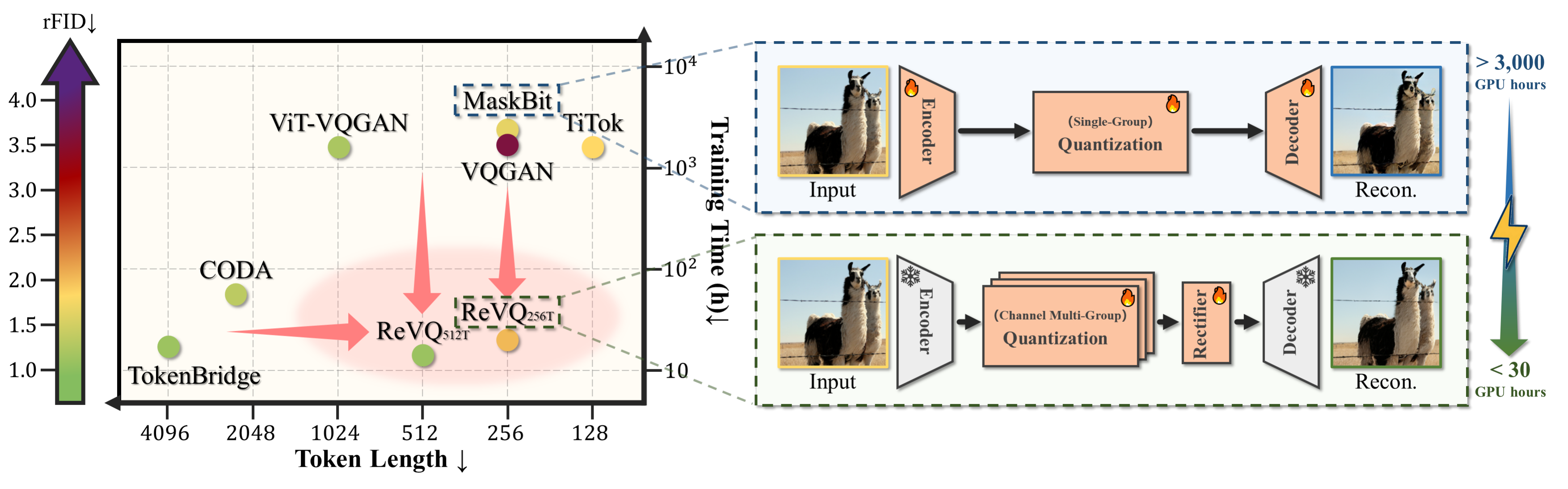}
    \caption{ReVQ achieves the optimal trade-off between training efficiency (1 day on a single NVIDIA 4090) and compression ratio ($\leq 512$ tokens for $256\times 256$ images), maintaining competitive reconstruction quality compared to SOTA VQ-VAEs like MaskBit~\citep{weber2024maskbit} ($4.5$ day on 32 A100s).}
    \label{fig:head}
    \vspace{-2mm}
\end{figure}

Large language models (LLMs)~\citep{brown2020language} have revolutionized artificial intelligence by utilizing discrete token sequences for next-token prediction. 
For integrating vision with LLMs, visual tokenizers play a critical role in bridging continuous image spaces and discrete input formats of LLMs. 
Vector-quantized variational autoencoders (VQ-VAEs)~\citep{van2017neural} serve as foundational components for this integration by discretizing image latent spaces, enabling alignment between visual and linguistic modalities in vision-LLM architectures~\citep{esser2021taming,razavi2019generating}.

Despite advancements in reconstruction quality~\citep{yu2022vector,chang2022maskgit}, modern VQ-VAEs face a fundamental challenge: a trade-off between \textbf{training efficiency} and \textbf{compression ratio}. 
Current approaches can be categorized into two distinct categories (\cref{fig:head}). 
(1) \textbf{high-compression but high-cost methods} (e.g., MaskBit~\citep{weber2024maskbit}, $\leq 256$ tokens) demand substantial computational resources, requiring over $3,000$ gpu hours on A100 clusters. 
This limits accessibility to well-resourced institutions. 
(2) \textbf{efficient but low-compression methods} (e.g., TokenBridge~\citep{wang2025bridging}, 4096 tokens; CODA~\citep{liu2025coda}, 2560 tokens) leverage pre-trained VAEs for rapid quantization but fail to achieve the short token lengths necessary for downstream generative tasks~\citep{rombach2022high}.

This work addresses the unmet need for a VQ-VAE framework that concurrently achieves high compression ratios and efficient training. 
We uncover an inherent relationship between VAEs and VQ-VAEs: under specific conditions, a pre-trained VAE can be systematically transformed into a VQ-VAE with minimal computational overhead. 
Unlike previous attempts such as TokenBridge~\citep{wang2025bridging} and CODA~\citep{liu2025coda}, which compromise on token length, our \textbf{Quantize-then-Rectify (ReVQ)} framework leverages pre-trained VAEs to facilitate fast VQ-VAE training while maintaining high compression performance (\cref{fig:framework}). 
By integrating \textbf{channel multi-group quantization} to expand codebook capacity and a \textbf{post rectifier} to alleviate quantization errors, ReVQ compresses ImageNet images into at most $512$ tokens while sustaining competitive reconstruction quality (rFID = $1.06$). 
ReVQ completes full training on a single NVIDIA 4090 in approximately $22$ hours, in contrast to comparable methods that require $4.5$ days on a 32 A100 GPUs. 
The core contributions of this work are as follows:
\begin{itemize}
    \item \textbf{Connection between VAE and VQ-VAE:} We formalize the boundary conditions for converting a VAE into a VQ-VAE, establishing a linkage between these two model classes.
    \item \textbf{Representative quantizer design:} A channel multi-group quantizer and a post rectifier are introduced to expand codebook capacity and mitigate quantization errors.   
    \item \textbf{Efficient ReVQ framework:} ReVQ enables converting a VAE into a VQ-VAE on a single NVIDIA 4090 within one day while ensuring reconstruction quality, achieving a two-order-of-magnitude improvement in training speed.
    \item \textbf{Extensive experimental analysis:} Results on ImageNet demonstrate ReVQ achieves superior balance between training efficiency and compression ratio, encoding images into $\leq 512$ tokens with competitive rFID while drastically reducing computational demands.
\end{itemize}

%% file: chapters/2_related_work.tex
\section{Related Work}

VQ-VAEs~\citep{van2017neural} bridge continuous and discrete spaces, enabling the application of deep learning in diverse domains such as image understanding~\citep{baobeit2022,ge2024making,jinunified2024} and generation~\citep{esser2021taming,chang2022maskgit,tian2024visual}. 
Existing efforts to enhance VQ-VAEs can be broadly categorized into \textbf{model structure} and \textbf{quantization strategy}.

\paragraph{Model Structure.}
The original VQ-VAE~\citep{van2017neural} first introduced an effective framework for discretizing continuous data. 
However, early VQ-VAEs often suffered from suboptimal reconstruction quality. 
Subsequent research focused on refining model architectures to address this limitation. 
First, diverse backbone networks were developed to enhance model capacity. 
VQ-VAE2~\citep{razavi2019generating} employed a multi-scale quantization strategy to preserve high-frequency details, 
while integrating Vision Transformers~\citep{yu2022vector,yu2024image,cao2023efficient} significantly improved representational power. 
Second, the incorporation of generative adversarial networks (GANs)~\citep{goodfellow2014generative} brought substantial advancements.
VQGAN~\citep{esser2021taming} improved the perceptual quality of reconstructed images by combining GANs with perceptual loss functions~\citep{larsen2016autoencoding,johnson2016perceptual}. 
Third, semantic supervision emerged as an effective approach.
VAR~\citep{tian2024visual} utilized DINO~\citep{oquab2023dinov2} as a semantic prior to enhance reconstruction fidelity, 
while ImageFolder~\citep{liimagefolder} introduced a semantic branch in the quantization module supervised by contrastive loss.

\paragraph{Quantization Strategy.}
Conventional VQ-VAEs rely on nearest-neighbor search to map features to codebook entries, a method that has been shown to have limitations in optimization stability and codebook utilization. 
For improving optimization stability, various techniques have been proposed to enhance training robustness: low-dimensional codebooks~\citep{yu2022vector}, shared affine transformations~\citep{huh2023straightening,zhu2024scaling}, specialized initializations~\citep{huh2023straightening,zhu2024scaling}, and model distillation~\citep{yu2024image}. 
ViT-VQGAN~\citep{yu2022vector} observed the sparsity in high-dimensional feature spaces and demonstrated that reducing codebook dimensionality increases feature-code proximity, thereby improving code utilization. 
Shared affine transformations~\citep{huh2023straightening,zhu2024scaling} highlighted the sparsity and slowness of conventional codebook updates, proposing an affine layer to convert sparse updates into dense transformations for more efficient adaptation. 
K-Means initialization~\citep{huh2023straightening,zhu2024scaling} was identified as a reliable method to mitigate premature convergence from arbitrary initializations, while distillation from pre-trained models like MaskGiT~\citep{chang2022maskgit} further boosted performance~\citep{yu2024image}. 
OptVQ~\citep{zhang2024preventing} applied optimal transport theory to model the global distributional relationship between codes and features, enhancing matching accuracy. 
To expand codebook capacity, strategies such as residual mechanisms~\citep{lee2022autoregressive}, multi-head mechanisms~\citep{zheng2022movq}, and multi-group quantization~\citep{ma2025unitok} have been proposed. 
Meanwhile, lookup-free approaches like FSQ~\citep{mentzer2023finite} and LFQ~\citep{yulanguage} aimed to enhance reconstruction efficiency by avoiding explicit codebook lookups.

%% file: chapters/3_method.tex
\vspace{-2mm}
\section{Method}
\vspace{-2mm}

\begin{figure}[tbp]
    \centering
    \includegraphics[width=0.95\linewidth]{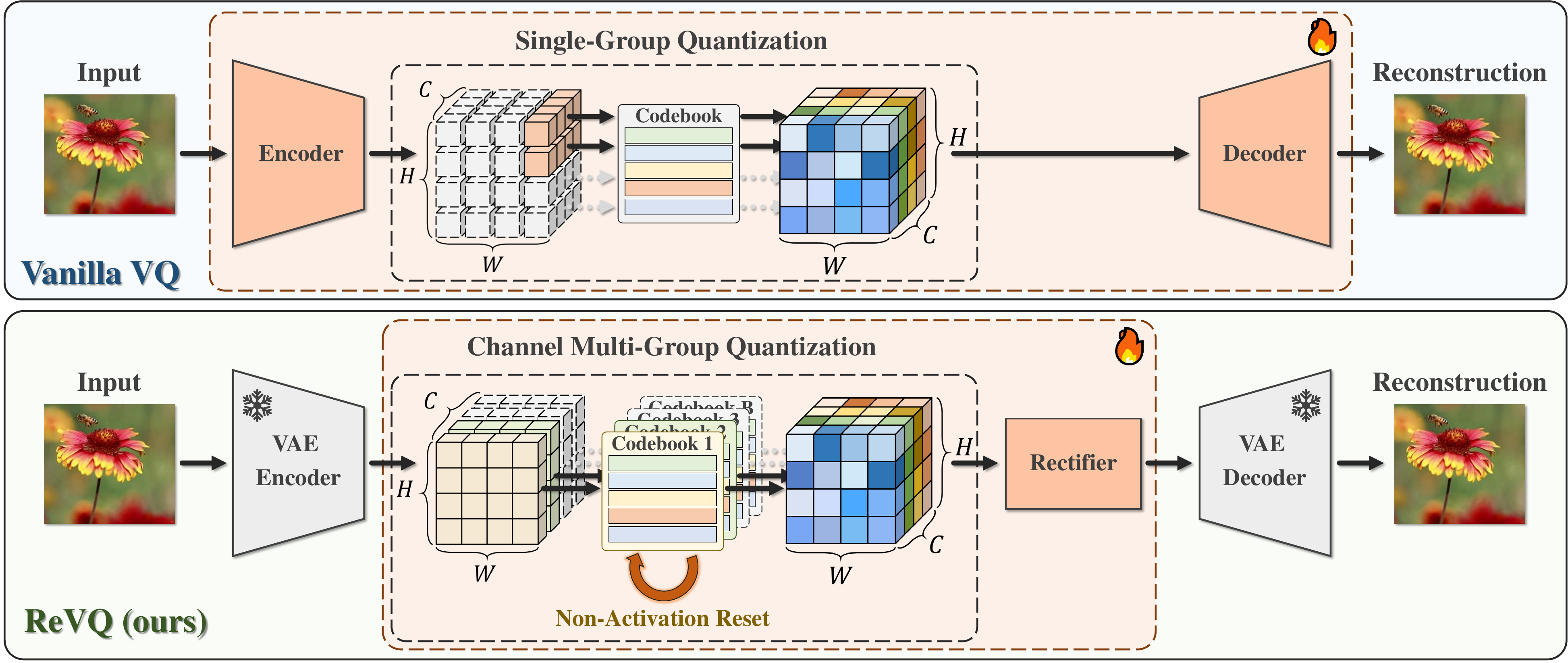}
    \caption{Comparison of Vanilla VQ and ReVQ.
    (Top) Vanilla VQ trains encoder, decoder, and quantizer from scratch, demanding substantial computational resources.
    (Bottom) ReVQ freezes pre-trained VAE encoder/decoder parameters, training only a quantizer and lightweight rectifier for high-performance VQ-VAE. To boost quantizer capacity, it uses channel multi-grouping and ensures codebook utilization via non-activation reset.}
    \label{fig:framework}
    \vspace{-3mm}
\end{figure}

Training a VQ-VAE from scratch is computationally expensive.
MaskBit~\citep{weber2024maskbit} reports 3456 GPU hours (4.5 days on a 32 A100 GPUs for 1.35M iterations) for ImageNet~\citep{deng2009imagenet}, which is prohibitive for most researchers.
This work addresses the high training cost by analyzing bottlenecks in VQ-VAEs and exploring strategies to accelerate VQ-VAE training.
In \cref{sec:vq-vae}, we dissect the core components of VQ-VAE and identify time-consuming modules.
\cref{sec:techniques} discusses key adaptations for converting VAEs to VQ-VAEs, including channel multi-group quantization and non-activation reset.
Finally, \cref{sec:revq} introduces ReVQ, a quantize-then-rectify approach that transforms pre-trained VAEs into VQ-VAEs with minimal computational overhead.

\vspace{-2mm}
\subsection{Preliminary: Time-Consuming VQ-VAE Training} \label{sec:vq-vae}
\vspace{-2mm}

Compact latent space representation of high-dimensional data is fundamental.
Autoencoders~\citep{hinton2006reducing} initiated the exploration of low-dimensional image encoding, while VAEs~\citep{kingma2013auto} advanced this by introducing prior distributions, enabling data generation via latent space sampling. 
With the GPT era, discrete image representations became necessary to align with discrete base LLMs. 
VQ-VAEs~\citep{van2017neural} replaced continuous priors with discrete codebooks, gaining wide use in image generation~\citep{esser2021taming,chang2022maskgit,rombach2022high} and large-scale pre-training~\citep{baobeit2022,bai2024sequential}. 
Let $\mathcal{X} = \{\vx_i\}_{i=1}^N$ denote the image dataset. 
A standard VQ-VAE consists of an encoder $f_e(\cdot)$, a decoder $f_d(\cdot)$, and a quantizer $q(\cdot)$. 
The encoder maps input image $\vx$ to a 3D latent feature $\mZ_e = f(\vx) \in \mathbb{R}^{H\times W\times D}$. 
For each vector $\vz_e \in \mathbb{R}^{D}$ in $\mZ_e$, the quantizer finds the nearest code vector in codebook $\mathcal{C} = \{\vc_1, \vc_2, \ldots, \vc_n\}$ via nearest-neighbor search, yielding the quantized vector $\vz_q = q(\vz_e)$. 
These form the quantized feature map $\mZ_q$, from which the decoder reconstructs the image as $\hat{\vx} = f_d(\mZ_q)$.

\paragraph{Training Pipeline.}
The quantizer is typically implemented via nearest neighbor search~\citep{van2017neural,esser2021taming} as:
\begin{align} \label{equ:nearest}
    \vz_q = q(\vz_e, \mathcal{C}) = \vc_k, ~ \text{where} ~ k = \argmin_{j} \Vert \vz_e - \vc_j \Vert,
\end{align}
with $\Vert \cdot \Vert$ denoting a distance metric (e.g., Euclidean). 
Large codebooks incur significant computational costs for distance matrix computation. 
Lookup-free quantizers~\citep{mentzer2023finite,yulanguage} avoid this by directly rounding feature map elements to integers for codebook indices.
However, quantization operation may get trapped in local minima, causing "index collapse"~\citep{huh2023straightening}. 
To address this, some replace nearest neighbor search with distribution matching~\citep{zhang2024preventing} to ensure full codebook utilization.
Existing works train VQ-VAEs end-to-end by minimizing reconstruction loss $\mathcal{L}_{\text{rec}} = \Vert \vx - \hat{\vx} \Vert$~\citep{van2017neural}. 
Since reconstruction may overemphasize low-level details, perceptual and adversarial losses~\citep{esser2021taming,chang2022maskgit,yu2022vector,cao2023efficient} are often added to enhance visual quality. 
Adversarial loss has the most significant aesthetic impact, followed by perceptual loss, $l_1$-based, and $l_2$-based reconstruction losses. 
The non-differentiable nearest neighbor search requires gradient approximation via the straight-through estimator~\citep{bengio2013estimating,huh2023straightening}.

\begin{figure}[tbp]
    \centering
    \subfloat[\label{fig:flops}Computation Statistics.]{
        \includegraphics[width=0.27\linewidth]{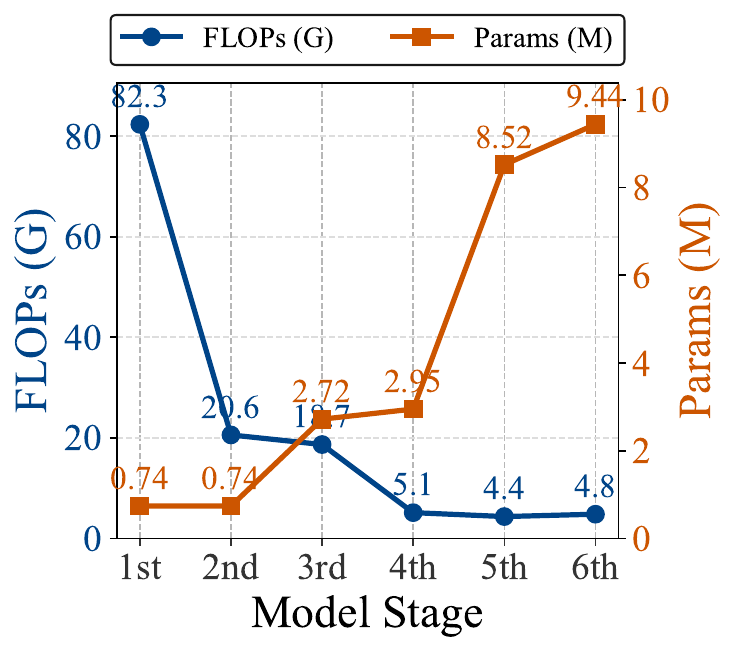}
    }
    \subfloat[\label{fig:noise_tolerent}Anlysis of Noise tolerant ability of the VAE under different noise levels.]{
        \includegraphics[width=0.7\linewidth]{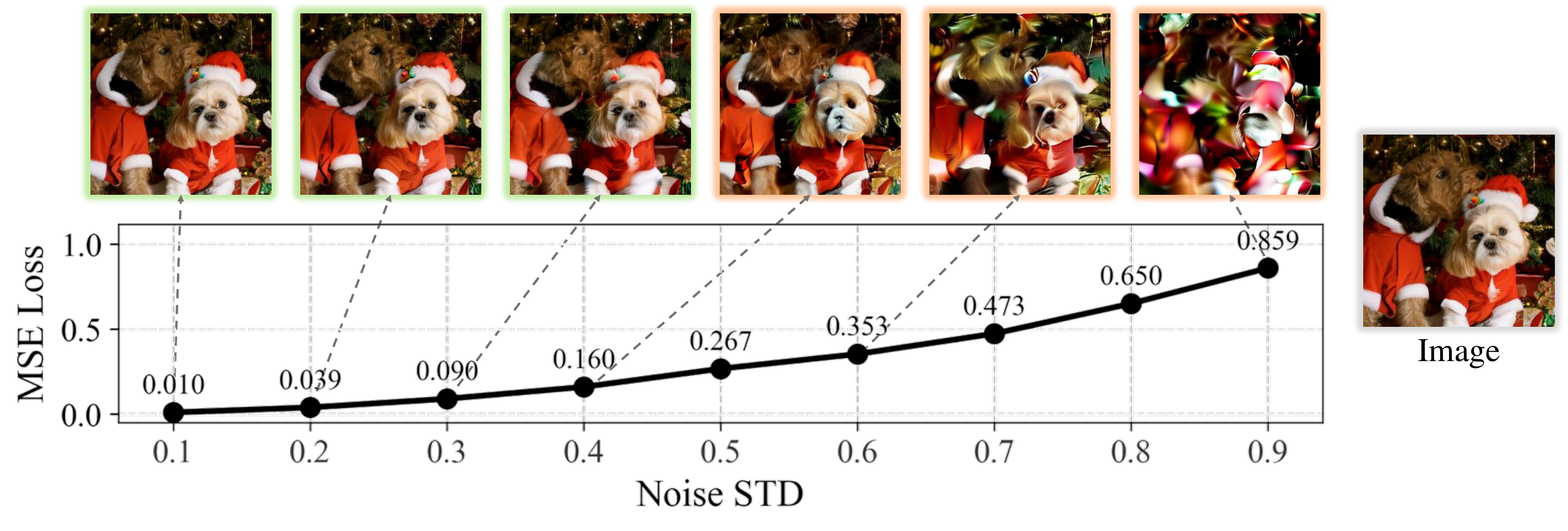}
    }
    \caption{Research Motivation. (a) Computational statistics reveal that shallow layers dominate computations, enabling substantial savings via pre-trained VAE. (b) VAE noise tolerance is demonstrated, showing conversion to VQ-VAE feasible when quantization error is below the threshold.}
    \vspace{-3mm}
\end{figure}

\paragraph{Computation Statistics.}
To understand why VQ-VAE training is time-consuming, we analyze FLOPs and parameters of a typical model~\citep{weber2024maskbit} (see \cref{fig:flops}). 
High computational burden concentrates in shallow layers due to large input resolution, while deep layers have more parameters. 
This motivates training VQ-VAEs on pre-downscaled inputs using pre-trained VAEs to compress images into latent spaces first. 
Recent works like TokenBridge~\citep{wang2025bridging} (4096 tokens per image) and CODA~\citep{liu2025coda} (2560 tokens per image via residual coding) demonstrate fast VQ-VAE development with pre-trained VAEs, though achieving lower compression ratios than conventional models (256 tokens per image). 
Our work explores whether fine-tuning pre-trained VAEs can yield VQ-VAEs with comparable compression efficiency and fast training.

\subsection{Convert VAE into VQ-VAE} \label{sec:techniques}

In this section, we observe strong noise tolerance in autoencoders and present key techniques for converting a VAE to a VQ-VAE. 
We encode an image from ImageNet by DC-AE~\citep{chen2024deep} to a 2048D latent vector, normalize it via dataset statistics, and add Gaussian noise with varying variances before decoder reconstruction (\cref{fig:noise_tolerent}). 
Results show reconstructed images retained high visual quality when noise variance $\leq 0.3$ (green boxes) but degraded significantly above this threshold (red boxes). 
This indicates that while \cref{equ:nearest}-based vector quantization introduces noise, acceptable reconstruction quality is maintained if quantization noise remains within the VAE's tolerance threshold. 
Critical factors include the \textbf{codebook capacity} and \textbf{optimization stability} to avoid local minima.

\subsubsection{Codebook Capacity: Channel Multi-Group Quantization} \label{sec:multi_channel}

The effective codebook capacity is critical for achieving low quantization error. 
Consider a sample encoded with $B$ tokens from a codebook of size $N$.
The number of token combinations is $M = N^B$, defining the upper bound of samples the codebook can represent. 
In practice, VQ-VAE's effective codebook capacity $M$ often far exceeds training data size.
For instance, ImageNet has $A = 1,281,167 \approx 2^{20.29}$ images, and a VQ-VAE with $B = 256$ tokens and $N = 1024$ yields $M = 1024^{256} \approx 2^{2560}$, where $M \gg A$.
This raises the question: why does such vast capacity not alleviate VQ-VAE training difficulty?
The answer lies in the shared codebook of conventional quantizers with spatial dimension spliting. 
For an encoded feature map $\mZ_e \in \mathbb{R}^{H\times W\times D}$, merging the first two spatial dimensions results in $\mZ'_e \in \mathbb{R}^{S \times D}$, where encoding length is $B = S = H \times W$. 
Traditional VQ uses a single codebook $\mathcal{C}$: each spatial location's feature vector $\vz_e^i = [\mZ'_e]_{(\cdot,i)}$ is quantized as $\vz_q^i = q(\vz_e^i, \mathcal{C})$. 
This imposes strong symmetry inductive bias by assuming identical prior distributions $p(\vz_e^i)$, limiting practical degrees of freedom to $N$ and causing training challenges, as demonstrated in the 2D example of \cref{fig:exp_multi_group} in \cref{sec:ablation_vq}.

\begin{figure}[tbp]
    \centering
    \subfloat[\label{fig:channel_multi_group}Comparison of different quantization strategies.]{
        \includegraphics[width=0.54\linewidth]{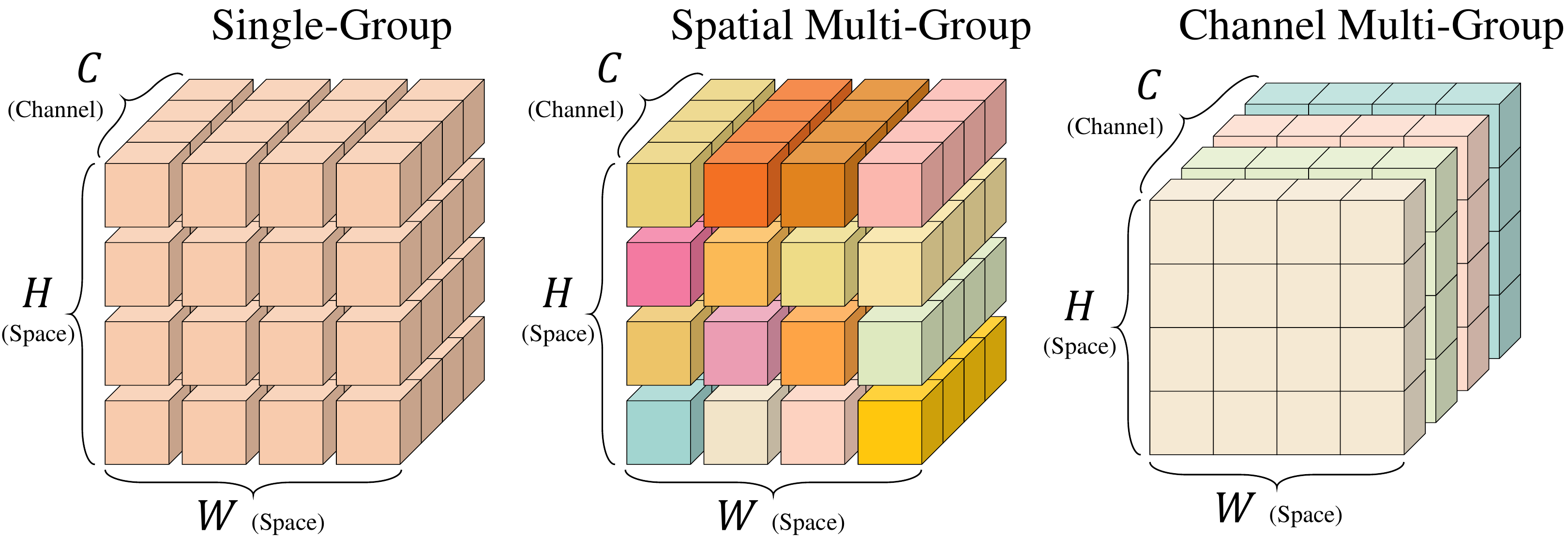}
    }
    \subfloat[\label{fig:channel_space}Correlation of channel and spatial features.]{
        \includegraphics[width=0.42\linewidth]{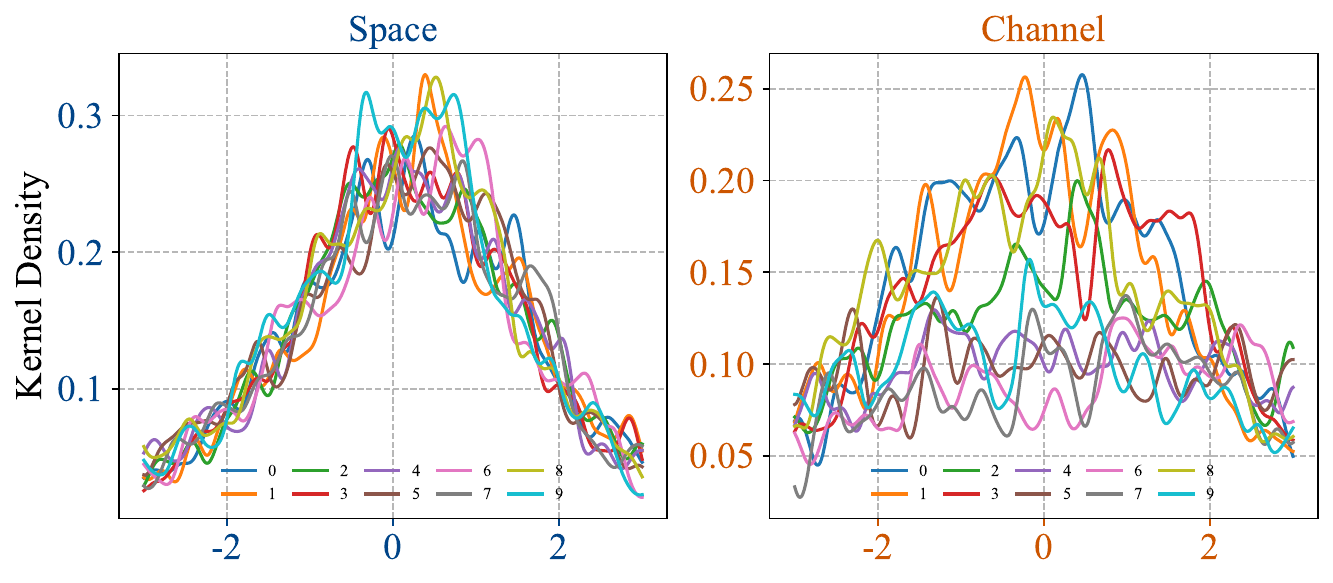}
    }
    \caption{(a) Three different quantizers. (b) Shows feature correlations under spatial/channel spliting, with channel-based spliting yielding more independent distributions.}
    \vspace{-3mm}
\end{figure}

To address this, multi-group strategies \citep{ma2025unitok} introduce $B$ independent codebooks $\mathcal{C}_i$, quantizing as 
\begin{align} \label{equ:multi_group}
    \vz_q^i = q(\vz_e^i, \mathcal{C}_i), ~i=1,\cdots,B.
\end{align}
If divided by spacial axis, then $B=S$.
This increases effective degrees of freedom to $N \times B$, mitigating training difficulties. 
Our research finds spatial spliting suboptimal for visual data.
\cref{fig:channel_space} shows kernel density statistics via linear dimensionality reduction on feature maps partitioned by spatial vs. channel dimensions.
This reveals highly correlated $\vz_e^i$ distributions under spatial spliting, versus relatively independent distributions under channel spliting. 
To fully utilize the flexibility of multi-group strategy, we propose a \textbf{channel multi-group} strategy: define $\vz_e^i = [\mZ'_e]_{(i,\cdot)}$ and apply multi-group quantization. 
A comparative illustration of the three methods is shown in \cref{fig:channel_multi_group} and \cref{tab:channel_multi_group}.
When token length $B$ differs from feature dimension $D$, we perform secondary spatial spliting after initial channel-wise division, resulting in feature vectors of dimension $d = (H \times W \times D)/B$.

\begin{table}[htbp] \small
    \vspace{-2mm}
    \centering
    \caption{Comparison of different quantizers.}
    \label{tab:channel_multi_group}
    \begin{tabular}{c|cccc}
        \hline
        \textbf{Strategy} & \textbf{\#Codebooks} & \textbf{Degrees of Freedom} & \textbf{Split Axis} & \textbf{Feature Diversity} \\
        \hline
        Single-Group & 1 & $N$ & Space: $[\mZ'_e]_{(\cdot,i)}$ & \scriptsize{\XSolidBrush} \\
        Spatial Multi-Group & $B$ & $N \times B$ & Space: $[\mZ'_e]_{(\cdot,i)}$ & \scriptsize{\XSolidBrush} \\
        Channel Multi-Group & $B$ & $N \times B$ & Channel: $[\mZ'_e]_{(i,\cdot)}$ & \scriptsize{\Checkmark} \\
        \hline
    \end{tabular}
    \vspace{-2mm}
\end{table}

\subsubsection{Optimization Stability: Non-Activation Reset} \label{sec:method_reset}

Extensive research has shown that optimizing VQ-VAE is a highly non-convex problem, extremely prone to index collapse~\citep{huh2023straightening,zhang2024preventing,zhu2024scaling}.
Failing to address its unstable optimization may render increased codebook capacity ineffective. 
Some studies suggest that alternating K-means initialization can resolve the unstable optimization~\citep{huh2023straightening}. 
Nevertheless, in large-scale problems, the overhead of K-means initialization is relatively high.
OptVQ~\citep{zhang2024preventing} points out the fundamental reason why nearest-neighbor based quantizers are prone to local optima.
Once a code $\vc_i$ is not selected by any sample in an iteration, it is highly likely to never be selected thereafter. 
Inspired by this, we identify that the key to solving this problem lies in "discovering non-activations and resetting them", thus proposing the \textbf{Non-Activation Reset} strategy. 
Specifically, during each training epoch, for the codebook $\mathcal{C}$, we count the activation time $t_i$ of each code $\vc_i$. 
At the end of the epoch, we sort the indices of the $N$ codes in ascending order of their $t_i$ values, obtaining $\sI = \{i_1, i_2, \cdots, i_N\}$. 
When there are $r$ unactivated codes (i.e., the first $r$ indices in $\sI$ have $t_i = 0$), we perform the following reset operation:
\begin{align} \label{equ:reset}
    \vc_{i_u} \leftarrow \vc_{i_{N + 1 - u}} + \epsilon, ~ u=1, \cdots, r,
\end{align}
where $\epsilon$ is a small random perturbation to avoid overlapping between codes after reset. 
This operation intuitively resets unactivated points to the vicinity of highly activated codes, 
sharing the burden of frequently activated codes and promoting a more uniform activation frequency across codes.

We find that methods balancing codebook activation frequencies effectively prevent codebook collapse.
This is also reflected in the entropy regularization proposed in LFQ~\citep{yulanguage} and the optimal transport search in OptVQ~\citep{zhang2024preventing}. 
The distinction lies in that our reset strategy requires no additional loss functions or computational steps during training.
Only a single reset operation at the end of each epoch, making it a plug-and-play module in code implementation.
We briefly analyze in the supplementary material (\cref{sec:theo_reset_analysis}) whether the reset strategy can reduce quantization errors.

\subsection{ReVQ: Quantize-then-Rectify} \label{sec:revq}

We primarily analyzed the essential components for converting a VAE into a VQ-VAE purely from the quantizer perspective. 
However, as shown in \cref{fig:exp_rectify}, the VQ-VAE converted from DC-AE~\citep{chen2024deep} can at most compress images into $512$ tokens to achieve a moderately effective model if relying solely on the quantizer.
Further increasing the compression ratio would lead to an exponential explosion in the required number of codebook. 
To address this, we introduce the \textbf{Quantize-then-Rectify (ReVQ)} framework in this section.
The proposed method posits that for the quantized features $\mZ_q$ from quantizer $q$, a rectifier $g$ should be constructed. 
The reconstructed quantized features via the ReVQ method are thus given by:
\begin{align} \label{equ:revq}
    \mZ_e' = g\left(q(\mZ_e, \mathcal{C})\right).
\end{align}
Since the rectifier $g$ is trained under relatively low-resolution cases, the comparisons in \cref{tab:exp_training_time} demonstrate that ReVQ can convert a VAE into a VQ-VAE extremely efficiently on a single RTX 4090 GPU. 
In contrast, traditional VQ-VAE training may require 4.5 days on 32 A100 GPUs as reported in \citep{weber2024maskbit}. 
We now elaborate on the rectifier model design and training loss function.

\paragraph{Rectifier Design.}
DC-AE~\citep{chen2024deep} is a highly practical study that proposes a high-compression VAE architecture capable of compressing images into $2048D$ vectors. 
This model employs a specially designed residual structure for image reconstruction and incorporates EfficientViT blocks~\citep{cai2023efficientvit} in deeper stages. 
In our ReVQ framework, since we do not involve upsampling/downsampling of latent variables, we directly utilize an EfficientViT block as the rectifier model $g$, which maintains consistent input and output dimensions.

\paragraph{Training Loss.}
Conventional VQ-VAE training typically involves a combination of loss functions, such as perceptual loss~\citep{johnson2016perceptual}, Patch GAN loss~\citep{isola2017image}, and standard $l_2$/ $l_1$ losses. 
In our ReVQ framework, however, to avoid heavy computational loads, we treat the VAE as a black box without computing its gradients. 
Consequently, we only apply $l_2$ loss in the latent space of $\mZ_e$ for training. 
The final optimization objective is:
\begin{align} \label{equ:loss}
    \min_{\theta_g, \mathcal{C}} L_{\text{ReVQ}} = \left\Vert \mZ_e - g\left(q(\mZ_e)\right) \right\Vert^2_2,
\end{align}
where $\theta_g$ denotes the parameters of the rectifier model and $\mathcal{C}$ represents all codebook parameters. 
The detailed training algorithm for ReVQ is provided in \cref{alg:revq}.

%% file: chapters/4_experiment.tex
\section{Experiment} \label{sec:exp}

In this section, we illustrate the reconstruction performance and training efficiency of ReVQ. 
We first detail the experimental setup in \cref{sec:exp_detail}, followed by presenting quantitative and qualitative comparisons between ReVQ and other VQ-VAE methods in \cref{sec:comparison}. 
Subsequently, ablation experiments are conducted in \cref{sec:ablation_vq} and \cref{sec:ablation_recti} to validate the effectiveness of the quantization module and the rectification module, respectively.

\begin{table}[tbp] \small
  \centering
  \caption{Quantitative comparison with state-of-the-art methods on ImageNet.}
  \label{tab:imagenet-recon}
  \begin{tabular}{|l|l|l|c|cccc|}
  \hline
  \textbf{Type} & \textbf{Method} & \textbf{Token Length} & \textbf{\#Codebook} & \textbf{SSIM↑} & \textbf{PSNR↑} & \textbf{LPIPS↓} & \textbf{rFID↓} \bigstrut\\
  \hline
  \multirow{10}[2]{*}{\makecell{From \\ Scratch}} & ViT-VQGAN~\citep{yu2022vector} & 1024 \tiny{(16$\times$16)} & 8,192 & -     & -     & -     & 1.28  \bigstrut[t]\\
        & Mo-VQGAN~\citep{zheng2022movq} & 1024 \tiny{(16$\times$16$\times$4)} & 1,024 & 0.673  & 22.420  & 0.113  & 1.12  \\
        & ImageFolder~\citep{liimagefolder} & 572 \tiny{(286$\times$2)} & 4,096 & -     & -     & -     & 0.80  \\
        & VQGAN~\citep{esser2021taming} & 256 \tiny{(16$\times$16)} & 16,384 & 0.542  & 19.930  & 0.177  & 3.64  \\
        & MaskGIT~\citep{chang2022maskgit} & 256 \tiny{(16$\times$16)} & 1,024 & -     & -     & -     & 2.28  \\
        & RQ-VAE~\citep{lee2022autoregressive} & 256 \tiny{(8$\times$8$\times$4)} & 16,384 & -     & -     & -     & 3.20  \\
        & MaskBit~\citep{weber2024maskbit} & 256 \tiny{(16$\times$16)} & 4,096 & -     & -     & -     & 1.61  \\
        & COSMOS~\citep{agarwal2025cosmos} & 256 \tiny{(16$\times$16)} & 64,000 & 0.518  & 20.490  & -     & 2.52  \\
        & VQGAN-LC~\citep{zhu2024scaling} & 256 \tiny{(16$\times$16)} & 100,000 & 0.589  & 23.800  & 0.120  & 2.62  \\
        & LlamaGen-L~\citep{sun2024autoregressive} & 256 \tiny{(16$\times$16)} & 16,384 & 0.675  & 20.790  & -     & 2.19  \bigstrut[b]\\
  \hline
  \multirow{2}[2]{*}{\makecell{Fine \\ Tuning}} & TiTok-S-128~\citep{yu2024image} & 128   & 4,096 & -     & -     & -     & 1.71  \bigstrut[t]\\
        & CODA~\citep{liu2025coda} & 2560 \tiny{(256$\times$10)} & 65,536 & 0.602  & 22.200  & -     & 1.34  \bigstrut[b]\\
  \hline
  \multirow{4}[2]{*}{Frozen} & TokenBridge~\citep{wang2025bridging} & 4096  & 64    & -     & -     & -     & 1.11  \bigstrut[t]\\
        & \textbf{ReVQ\textsubscript{512T}} & \textbf{512}   & \textbf{16,384} & \textbf{0.690}  & \textbf{23.700}  & \textbf{0.092}  & \textbf{1.06}  \\
        & \textbf{ReVQ\textsubscript{256T}} & \textbf{256}   & \textbf{65,536} & \textbf{0.620}  & \textbf{21.690}  & \textbf{0.129}  & \textbf{2.57}  \\
        & \textbf{ReVQ\textsubscript{256T}} & \textbf{256}   & \textbf{262,144} & \textbf{0.640}  & \textbf{21.960}  & \textbf{0.121}  & \textbf{2.05}  \bigstrut[b]\\
  \hline
  \end{tabular}
  \vspace{-3mm}
\end{table}

\begin{figure}[tbp]
    \centering
    \includegraphics[width=0.85\linewidth]{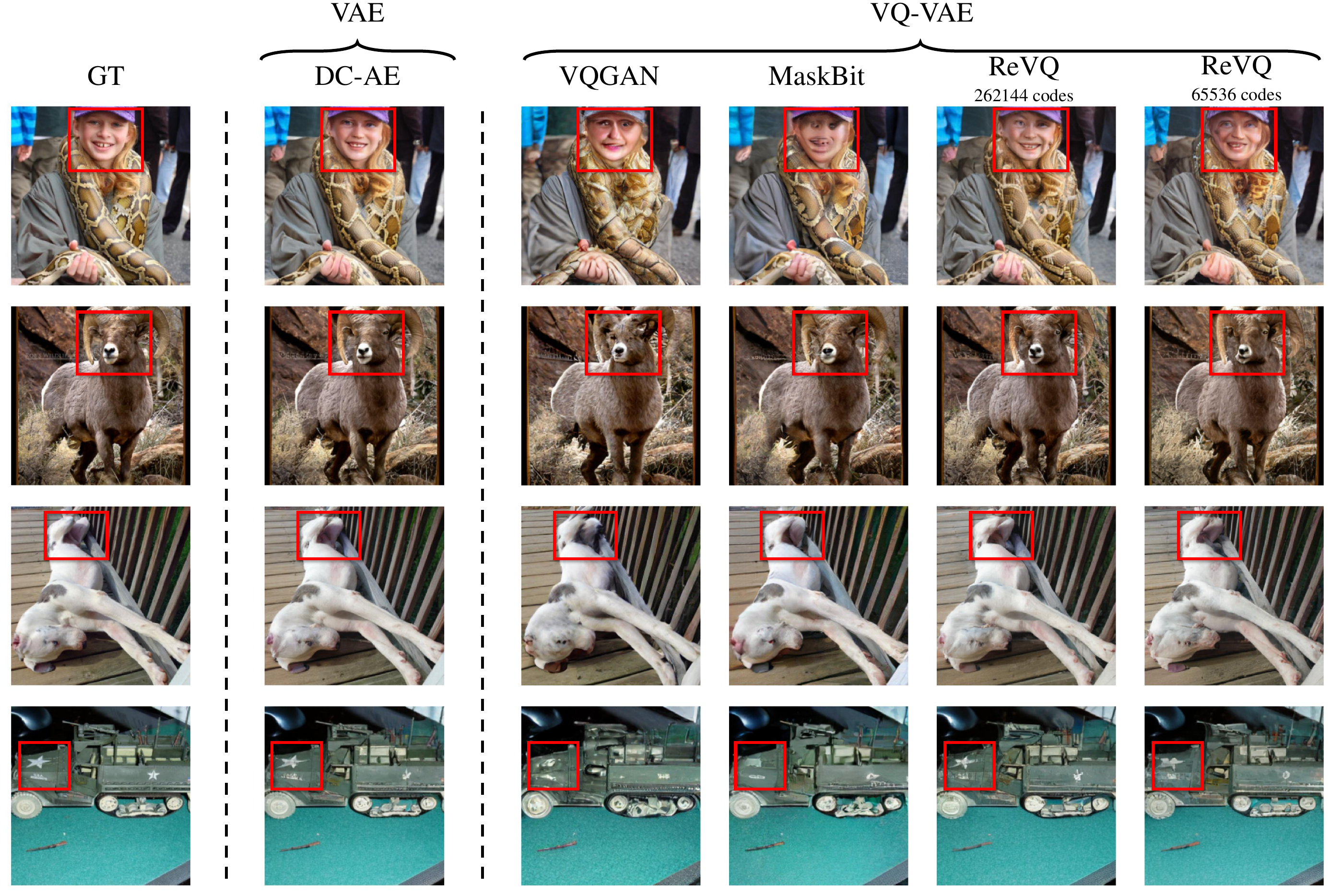}
    \caption{Reconstruction results on ImageNet validation set (details marked in red boxes).}
    \label{fig:recon}
    \vspace{-4mm}
\end{figure}

\subsection{Experimental Detail} \label{sec:exp_detail}

\paragraph{Model Setting.}
Our model consists of a quantizer and a decoder. 
Building upon the continuous latent space of the VAE, 
we utilize and freeze both the encoder and decoder weights from the DC-AE~\citep{chen2024deep} without further modification.
ImageNet~\citep{deng2009imagenet} images are first encoded via the VAE encoder,
after which the resulting features are normalized using the global mean and variance computed from the entire dataset. 
These normalized features are then provided as input to our model and subsequently compressed into tokens of either 512 or 256 dimensions.
The model is optimized by minimizing the $\ell_2$ loss between the normalized latent features and their reconstructions.
The quantizer incorporates a \textbf{channel multi-group} strategy in conjunction with a \textbf{non-activation reset} strategy.
The codebook size is 16384 for 512 token length, while it is set to 65536 or 262144 for 256 token length.
Unless otherwise specified, we implement the rectifier using a three-layer EfficientViT~\citep{cai2023efficientvit} block for 512 token length and a four-layer EfficientViT block for 256 token length.
Both configurations maintain consistent input and output dimensions.

\paragraph{Optimizer Setting.}
All models are implemented using the PyTorch~\citep{paszke2019pytorch} framework and trained on a single NVIDIA 4090 GPU.
AdamW~\citep{loshchilov2019decoupled} functions as the optimizer throughout.
The batch size is determined dynamically based on the codebook size and GPU memory constraints.
Specifically, when the token length is configured as 512, a batch size of 128 is employed.
For a token length of 256, a batch size of 256 is applied with a codebook size of 65536,  
and reduced to 128 when the codebook size increases to 262144. 
The learning rate for all quantizers is initialized at $1.0 \times 10^{-2}$, 
whereas the decoder's learning rate is fixed at $5\%$ of that of the quantizer.
An exponential learning rate scheduling policy is adopted.
All models underwent training for 100 epochs.

\subsection{Performance Comparison} \label{sec:comparison}

We conduct a comparative analysis of our ReVQ model against leading VQ-VAEs on ImageNet~\citep{deng2009imagenet}.
Evaluations are conducted on the validation, employing four standard metrics: PSNR, SSIM~\citep{wang2004image}, LPIPS~\citep{zhang2018unreasonable}, and rFID~\citep{heusel2017gans},
as summarized in \cref{tab:imagenet-recon}.
Two salient observations emerge from our results. 
First, the model with a token length of 512 demonstrates superior performance across all metrics, 
surpassing both ``Fine Tuning'' and ``Frozen'' counterparts. 
Additionally, the configuration with a token length of 256 and a codebook size of 262144 achieves notable outcomes, 
surpassing all other 256 token length models except MaskBit~\citep{weber2024maskbit}.
Second, our model exhibits a significant advantage in training efficiency. 
Compared with publicly available training durations of existing approaches, 
ReVQ reduces the total GPU hours by $40\times \sim 150\times$ in \cref{tab:exp_training_time}.
Furthermore, \cref{fig:recon} illustrates the visual reconstruction quality. 
The red-boxed regions highlight ReVQ’s superior ability to preserve fine-grained details, 
particularly in areas involving complex textures and facial features.

\begin{table}[tbp]
    \small
    \centering
    \begin{minipage}[t]{0.59\linewidth}
        \centering
        \caption{Training time across different methods.}
        \begin{tabular}{|c|c|c|c|}
        \hline
        \textbf{Method} & \textbf{\#Code} & \textbf{GPUs} & \textbf{Training Time} \bigstrut\\
        \hline
        MaskBit & 4,096 & 32$\times$A100 & 3456 \bigstrut[t]\\
        TiTok-S-128 & 4,096 & 32$\times$A100 & 1600 \bigstrut[b]\\
        \hline
        ReVQ\textsubscript{512T} & 16,384 & 1$\times$RTX 4090 & 22 \bigstrut[t]\\
        ReVQ\textsubscript{256T} & 65,536 & 1$\times$RTX 4090 & 26 \\
        ReVQ\textsubscript{512T} & 262,144 & 1$\times$RTX 4090 & 40 \bigstrut[b]\\
        \hline
        \end{tabular}
        \label{tab:exp_training_time}
    \end{minipage}
    \hfill
    \begin{minipage}[t]{0.39\linewidth}
        \centering
        \caption{Spatial/channel split.}
        \begin{tabular}{|c|c|c|c|}
        \hline
        \textbf{\#Token} & \textbf{\#Code} & \textbf{Type} & \textbf{rFID} \bigstrut\\
        \hline
        \multirow{2}[2]{*}{512} & \multirow{2}[2]{*}{16,384} & space & 1.11  \bigstrut[t]\\
            &       & channel & 1.06  \bigstrut[b]\\
        \hline
        \multirow{2}[2]{*}{256} & \multirow{2}[2]{*}{65,536} & space & 2.91  \bigstrut[t]\\
            &       & channel & 2.57  \bigstrut[b]\\
        \hline
        \end{tabular}
        \label{tab:exp_multi_type}
    \end{minipage}
    \vspace{-3mm}
\end{table}

\subsection{Ablation Study on Quantizer Design} \label{sec:ablation_vq}

\begin{figure}[bp]
    \vspace{-3mm}
    \centering
    \begin{minipage}[t]{0.49\linewidth}
        \centering
        \includegraphics[width=\linewidth]{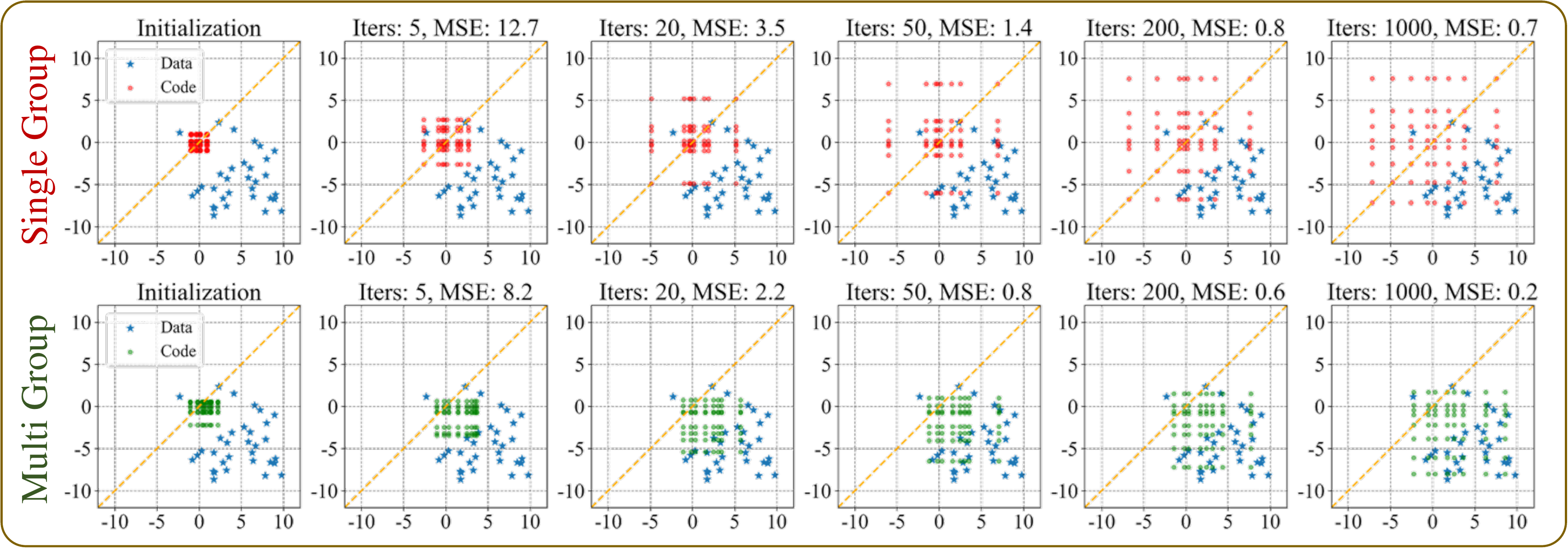}
        \caption{Single/multi-group strategy.}
        \label{fig:exp_multi_group}
    \end{minipage}
    \hfill
    \begin{minipage}[t]{0.49\linewidth}
        \centering
        \includegraphics[width=\linewidth]{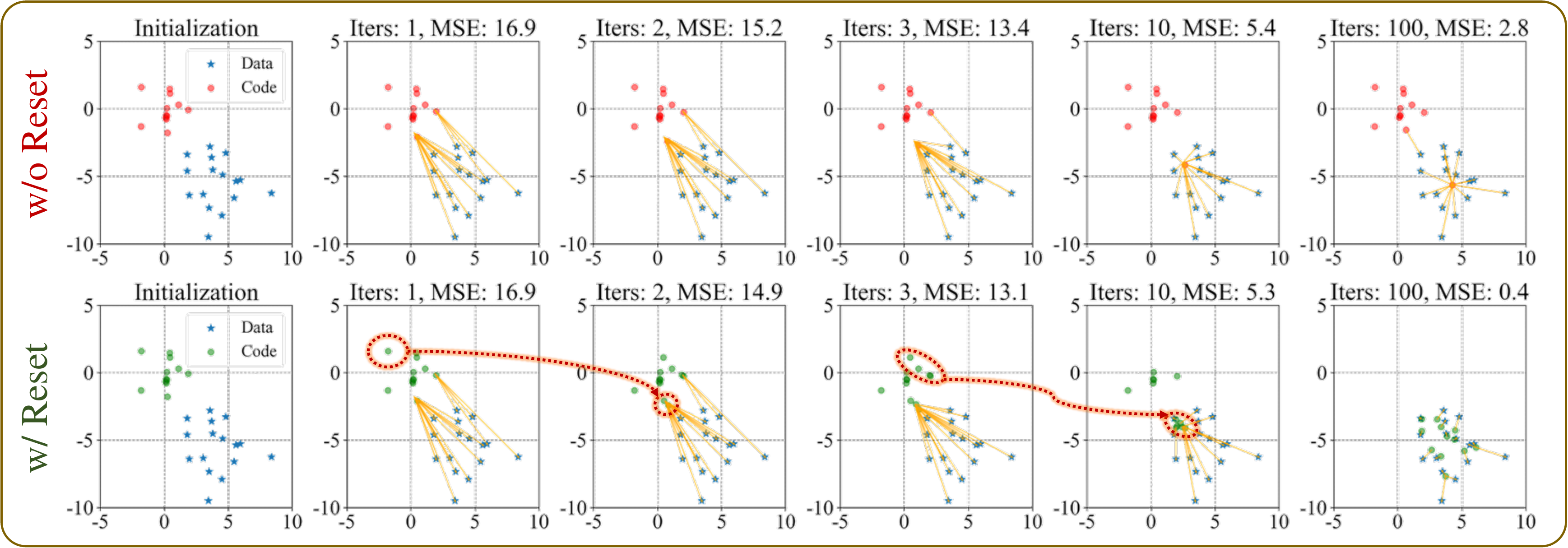}
        \caption{The influence of the reset strategy.}
        \label{fig:exp_reset}
    \end{minipage}

    \begin{minipage}[t]{0.56\linewidth}
        \centering
        \includegraphics[width=\linewidth]{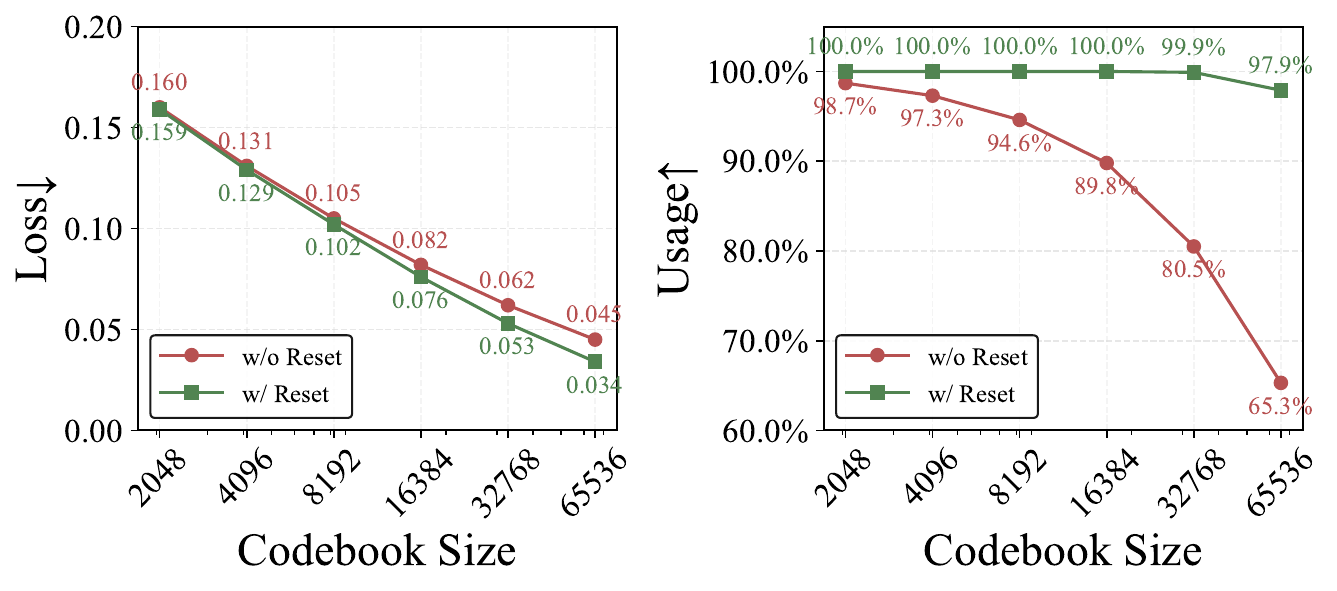}
        \caption{Ablation on reset strategy.}
        \label{fig:vq_reset}
    \end{minipage}
    \hfill
    \begin{minipage}[t]{0.42\linewidth}
        \centering
        \includegraphics[width=\linewidth]{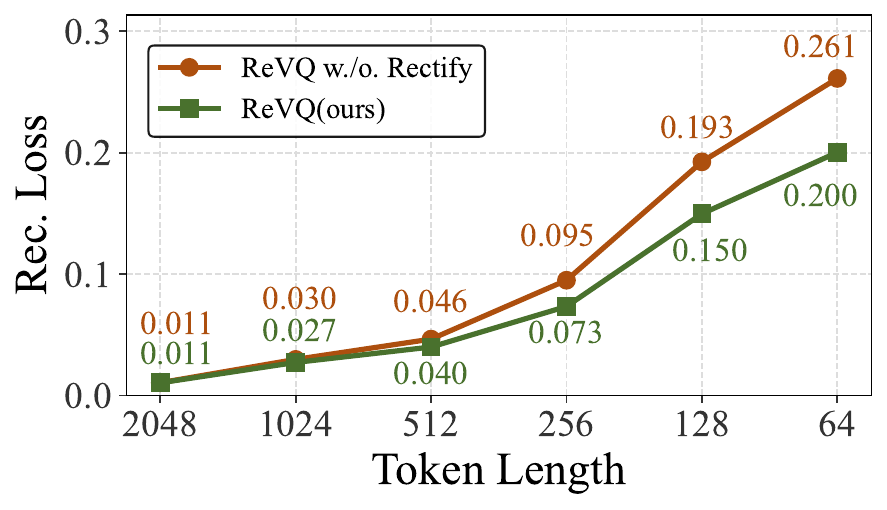}
        \caption{Ablation on rectifier.}
        \label{fig:exp_rectify}
    \end{minipage}
\end{figure}

\paragraph{Channel Multi-Group Strategy.}  
To ensure that the quantization error remains below the tolerance threshold of the frozen VAE model, we propose the \textbf{channel multi-group} strategy. 
\cref{fig:exp_multi_group} demonstrates the superiority of the multi-group strategy over the single-group strategy in quantization. 
We randomly initialized several 2D data points, with each data point represented by 2 tokens. 
For the single-group strategy, due to its inherent symmetry constraint, the reconstructed data points are forced to be symmetric about the line $y = x$, leading to a quantization error of $0.7$. 
In contrast, the multi-group strategy, with its higher degree of freedom, can better adapt to the true data distribution, achieving a minimum quantization error of $0.2$.  
We also quantitatively compared the performance of space-based and channel-based spliting. 
As shown in \cref{tab:exp_multi_type}, the rFID values of spliting along space and channel are presented respectively. 
It can be observed that under both 512-token and 256-token lengths, spliting along channel consistently outperforms space.

\paragraph{Non-Activation Reset Strategy.}  
Nearest-neighbor-based quantizers often face the challenge of codebook collapse~\citep{huh2023straightening,zhu2024scaling,zhang2024preventing}. 
To address this issue, we propose the \textbf{Non-Activation Reset} strategy in this paper. 
We first visualize the dynamic process of codebook changes under this strategy in \cref{fig:exp_reset}. 
We randomly initialized several 2D data points, each represented by 1 token. 
Without the reset strategy, the codebook is heavily influenced by the initialization, resulting in only a few codes being used (e.g., only 2 codes in this case) and a quantization error of $2.8$. 
With the Reset strategy, inactive codes are reset to data-dense regions during training, as shown by the orange dashed arrows in the figure. 
This ensures all codes are used, reducing the quantization error to $0.4$. 
To more thoroughly demonstrate the effectiveness of this strategy, we conducted quantitative experiments on 10\% of the ImageNet dataset, as shown in \cref{fig:vq_reset}. 
The results show that without the reset strategy, codebook utilization decreases rapidly as the codebook size increases, with only 65.3\% of the codes utilized. 
In contrast, with the reset strategy, codebook utilization remains above 97\% without significant decline as the codebook size increases.

\begin{table}[tbp]
    \small
    \centering
    \begin{minipage}[t]{0.49\linewidth}
        \centering
        \caption{Ablations on decoder types.}
        \begin{tabular}{|c|c|c|c|c|}
        \hline
        \textbf{\#Token} & \textbf{\#Code} & \textbf{Decoder} & \textbf{rFID} & \boldmath{}\textbf{$L_{\text{MSE}}$}\unboldmath{} \bigstrut\\
        \hline
        \multirow{3}[2]{*}{512} & \multirow{3}[2]{*}{16,384} & ViT   & 1.06  & 0.012  \bigstrut[t]\\
                &       & CNN   & 1.08  & 0.012  \\
                &       & MLP   & 1.09  & 0.013  \bigstrut[b]\\
        \hline
        \multirow{3}[2]{*}{256} & \multirow{3}[2]{*}{65,536} & ViT   & 2.57  & 0.064  \bigstrut[t]\\
                &       & CNN   & 3.00  & 0.068  \\
                &       & MLP   & 4.58  & 0.076  \bigstrut[b]\\
        \hline
        \end{tabular}
        \label{tab:exp_decoder_type}
    \end{minipage}
    \hfill
    \begin{minipage}[t]{0.49\linewidth}
        \centering
        \caption{Exploration of whether to add encoder.}
        \begin{tabular}{|c|c|c|c|c|}
        \hline
        \textbf{\#Token} & \textbf{\#Code} & \textbf{\makecell{with \\ Encoder?}} & \textbf{rFID} & \boldmath{}\textbf{$L_{\text{MSE}}$}\unboldmath{} \bigstrut\\
        \hline
        \multirow{2}[2]{*}{512} & \multirow{2}[2]{*}{16,384} & \scriptsize{\CheckmarkBold} & 189.10  & 0.620  \bigstrut[t]\\
            &       & \scriptsize{\XSolidBrush} & 1.06  & 0.012  \bigstrut[b]\\
        \hline
        \multirow{2}[2]{*}{256} & \multirow{2}[2]{*}{65,536} & \scriptsize{\CheckmarkBold} & 200.46  & 0.653  \bigstrut[t]\\
            &       & \scriptsize{\XSolidBrush} & 2.57  & 0.064  \bigstrut[b]\\
        \hline
        \end{tabular}
        \label{tab:exp_use_encoder}
    \end{minipage}
    \vspace{-3mm}
\end{table}

\subsection{Ablation Study on Rectifier Design} \label{sec:ablation_recti}

\paragraph{Effectiveness of Rectification.}
We initiate our analysis by evaluating the impact of the rectifier module on model performance in \cref{fig:exp_rectify}.
We conduct training on the ImageNet dataset using different token lengths and their corresponding codebook sizes,
with consistent training strategies and an identical rectifier design.
The use of the rectifier consistently reduces reconstruction loss across all token lengths. 
Notably, the improvement is more pronounced when the baseline model is weaker. 
Specifically, at a token length of 64, the rectifier yields a 23.3\% decrease in reconstruction loss,
highlighting its effectiveness in improving representational fidelity under constrained settings.

\paragraph{Diverse Rectifier Architectures.}
We further examine how different architectural designs of the rectifier affect model performance.
In particular, we investigate three rectifier designs employing ViT, CNN, and MLP backbones.
To rigorously evaluate the comparative performance of each rectifier design,
we conduct experiments on the ImageNet dataset under two settings:
one with a token length of 512 and a codebook size of 16384,
and another with a token length of 256 and a codebook size of 65536.
All other training settings are kept identical.
Empirical evidence indicates that the ViT rectifier consistently surpasses its CNN and MLP counterparts across both configurations.
Additionally, conventional VQ-VAEs adopt symmetrical architectural designs~\citep{agarwal2025cosmos,weber2024maskbit,yu2024image}. 
A natural question arises: why not use an additional encoder before the quantizer? 
In \cref{tab:exp_use_encoder}, we explored adding an encoder matching the rectifier's architecture before the quantizer to improve reconstruction performance. 
We found this greatly increased training difficulty, causing a significant rise in rFID. 
Thus, we ultimately chose not to add an extra encoder before the quantizer.

%% file: chapters/5_conclusion.tex
\section{Discussion} \label{sec:conclusion}

This paper addresses the issue of time-consuming training in conventional VQ-VAEs.
We discover that a pre-trained continuous feature autoencoder (VAE) and a discrete feature VQ-VAE exhibit an inherent connection. 
If the quantization error generated by the quantizer is smaller than the tolerance threshold, the VAE model can be seamlessly converted into a VQ-VAE model.  
Specifically, we propose a strategy named \textbf{Quantize-then-Rectify}. 
First, we freeze the parameters of the pre-trained VAE and directly apply a \textbf{channel multi-group quantization} strategy to transform continuous features into discrete tokens. 
During training, we introduce a simple \textbf{non-activation reset} strategy to address the commonly encountered ``codebook collapse'' problem. 
To further reduce quantization errors, a learnable ViT model is employed as a \textbf{rectifier} after the quantizer to correct the quantized tokens.  
Our experiments on the ImageNet dataset demonstrate that the proposed \textbf{ReVQ} method can achieve a VQ-VAE with high compression ratio after approximately 1 day of training on a single 4090 GPU server. 
In contrast, conventional VQ-VAE methods requiring comparable performance necessitate 4.5 days of training on a 32 A100 GPUs. 
However, as shown in \cref{fig:exp_rectify}, ReVQ currently cannot match state-of-the-art approaches like TiTok in achieving extremely high compression ratios (e.g., compressing images into 32 tokens). 
We attribute this limitation to the architectural design of the rectifier 
and plan to explore more reasonable model designs in future work to enhance the compression capability of ReVQ.  
Beyond this, we will investigate the applicability of ReVQ across more data modalities (such as video reconstruction) and downstream tasks (such as image generation), aiming to broaden the methodological scope of ReVQ.

%% file: chapters/appendix.tex
\newpage
\appendix

\startcontents[sections]
\printcontents[sections]{l}{1}{
    \setcounter{tocdepth}{2}
    \section*{\centering Table of Content for Appendix \rule{\linewidth}{0.5pt}}
}
\rule{\linewidth}{0.5pt}

\section{On the Theoretical Analysis of Reset Strategy} \label{sec:theo_reset_analysis}

We briefly analyze whether this Reset strategy can effectively reduce quantization error. 
Consider a code $\vc_1$ activated by $m$ feature vectors $\vz_i, i=1,\cdots,m$, with its quantization error given by:
\begin{align}
    L_{\text{MSE}}(\vc_1) = \sum_{i=1}^m \Vert \vz_i - \vc_1 \Vert^2_2.
\end{align}
Let $\bar{\vz}^{(m)}$ denote the mean of these $m$ vectors. By the least squares method, the quantization error reaches a lower bound $L^{(\text{lower})}_{m} \leq L_{\text{MSE}}(\vc_1)$ when $\vc_1 = \bar{\vz}^{(m)}$. 
If an unactivated code $\vc_2$ is reset near $\vc_1$, the $m$ feature vectors are divided into two subsets $\{\vz_i\}_{i=1}^{m_1}$ and $\{\vz_j\}_{j=1}^{m_2}$ with $m_1 + m_2 = m$. The updated quantization error becomes:
\begin{align}
    L'_{\text{MSE}}(\vc_1, \vc_2) = \sum_{i=1}^{m_1} \Vert \vz_i - \vc_1 \Vert^2_2 
    + \sum_{j=1}^{m_2} \Vert \vz_j - \vc_2 \Vert^2_2.
\end{align}
Similarly, the updated lower bound $L'^{(\text{lower})}_{m_1,m_2}$ is achieved when $\vc_1 = \bar{\vz}^{(m_1)}$ and $\vc_2 = \bar{\vz}^{(m_2)}$, satisfying:
\begin{align}
    &L'^{(\text{lower})}_{m_1,m_2} = \sum_{i=1}^{m_1} \Vert \vz_i - \bar{\vz}^{(m_1)} \Vert^2_2 
    + \sum_{j=1}^{m_2} \Vert \vz_j - \bar{\vz}^{(m_2)} \Vert^2_2 \notag \\
    \leq 
    &\sum_{i=1}^{m_1} \Vert \vz_i - \bar{\vz}^{(m)} \Vert^2_2 
    + \sum_{j=1}^{m_2} \Vert \vz_j - \bar{\vz}^{(m)} \Vert^2_2
    = L^{(\text{lower})}_{m}.
\end{align}
This analysis demonstrates that the Reset operation can effectively reduce quantization error. 
A 2D experiment in \cref{sec:ablation_vq} and the detailed in \cref{fig:exp_reset} illustrate how the Reset operation avoids codebook collapse and thereby decreases quantization error.

\section{Algorithm of ReVQ Method} \label{sec:revq_algorithm}

Below is the detailed algorithmic procedure for training one epoch using the ReVQ method.

\begin{algorithm}[htbp]
    \caption{One-epoch training algorithm for ReVQ.}
    \label{alg:revq}
    \begin{algorithmic}[1]
    \renewcommand{\algorithmicrequire}{\textbf{Input:}}
    \renewcommand{\algorithmicensure}{\textbf{Output:}}
    \REQUIRE Set of latent feature maps $\sZ_e$, quantizer $q$, rectifier $g$.
    \ENSURE  Optimized quantizer $q$ and rectifier $g$.
       \FOR {each $\mZ_e \in \sZ_e$}
          \STATE Quantize $\mZ_e$ to obtain $\mZ_q = q(\mZ_e, \mathcal{C})$ via channel multi-group strategy in \cref{sec:multi_channel}.
          \STATE Rectify $\mZ_q$ to produce $\mZ_q' = g(\mZ_q)$ via the rectifier model defined in \cref{sec:revq}.
          \STATE Calculate the loss function $L_{\text{ReVQ}}$ as specified in \cref{equ:loss}.
          \STATE Perform backpropagation to update the parameters of rectifier $g$ and codebook $\mathcal{C}$.
       \ENDFOR
       \STATE Apply Non-activation Reset to codebook $\mathcal{C}$ as defined in \cref{equ:reset}.
    \RETURN Quantizer $q$ and rectifier $g$.
    \end{algorithmic}
\end{algorithm}

\section{Implementation Details} \label{sec:appendix_implementation}

\subsection{Datasets} \label{sec:appendix_datasets}

This study was primarily conducted on the ImageNet dataset~\citep{deng2009imagenet}. 
The training set of ImageNet comprises 1281167 images, while the validation set contains 50000 images, both divided into 1000 classes. 
To enhance the training efficiency of the ReVQ model, we first employed the DC-AE model to encode all training images into 2048-dimensional vectors. 
The website for the ImageNet dataset is: \url{http://www.image-net.org/}.

\subsection{Configurations} \label{sec:appendix_config}

We did not employ any special data augmentation methods for the 2048-dimensional latent vectors. 
Taking our configuration with 512 tokens and a codebook size of 16384 as an example, 
the detailed settings for the model and the optimizer are as follows:
\begin{itemize}
    \item Num Code: 16384.
    \item Num Group: 512.
    \item Tokens Per Data: 512.
    \item Decoder: dc\_ae.
    \item In Channels: 32.
    \item Latent Channels: 32.
    \item Attention Head Dim: 32.
    \item Block Type: EfficientViTBlock.
    \item Block Out Channels: 512.
    \item Layers Per Block: 3.
    \item QKV Multiscales: [5].
    \item Norm Type: RMSNorm.
    \item Act Fn: SiLU.
    \item Upsample Block Type: interpolate.
    \item Optimizer: AdamW~\citep{loshchilov2019decoupled}.
    \item Beta1: 0.9.
    \item Beta2: 0.999.
    \item Quantizer Weight Decay: 0.0.
    \item Decoder Weight Decay: 1e-4.
    \item Learning Rate: 1e-4.
    \item LR Scheduler: ExponentialLR.
    \item Base LR: 1e-2.
    \item Epoch: 100.
    \item BatchSize: 256.
    \item GPU: One NVIDIA GeForce RTX 4090.
\end{itemize}

\section{Additional Experiments} \label{sec:appendix_experiments}

\subsection{Relationship between Token Length and Number of Codebooks} \label{sec:appendix_token_codebook}

\begin{figure}[htbp]
    \centering
    \includegraphics[width=0.95\linewidth]{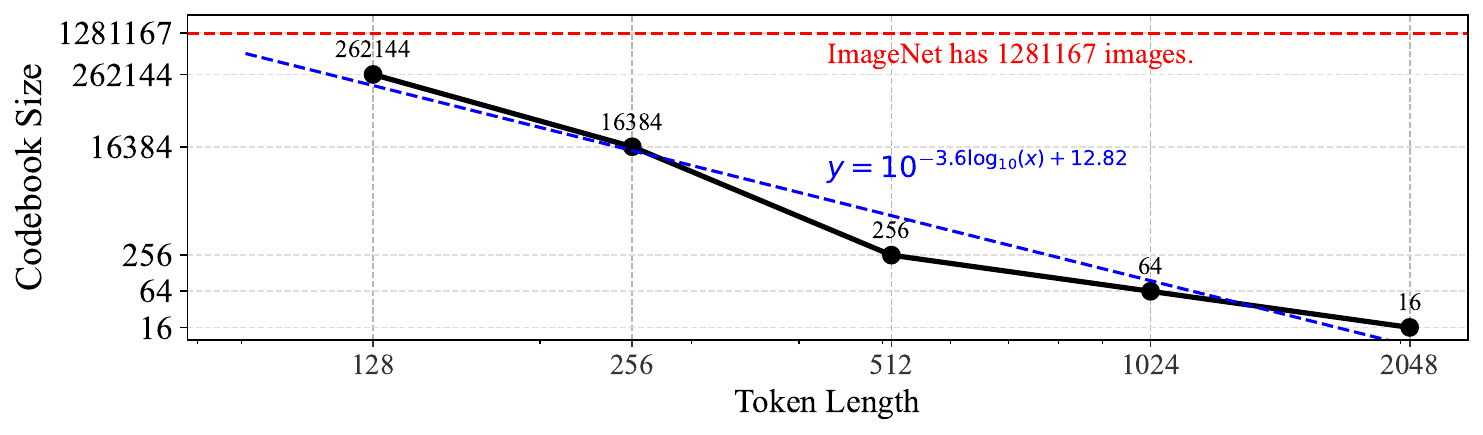}
    \caption{Relationship between the token length $B$ and the number of codebooks $M$ required to keep the quantization error below $0.1$.}
    \label{fig:appendix_fit_curve}
\end{figure}

We found that when the quantization error (MSE) of the latent vector is below $0.1$, 
the visual results of the reconstructed images are basically acceptable to the human eye. 
Since the dimension of the latent variable is $2048$, if the token length is $B$, the dimension of the codebook is $2048/B$. 
Obviously, a smaller token length $B$ leads to a higher codebook dimension. 
To ensure the quantization error is below $0.1$, a larger number of codebooks is required. 
Therefore, we explored the relationship between the token length $B$ and the minimum number of codebooks $M$ needed to keep the quantization error below $0.1$ in \cref{fig:appendix_fit_curve}. 
Specifically, only the quantizer was used in this experiment without employing the rectifier to further correct the quantization error.
It can be observed that the minimum number of codebooks $M$ and the token length $B$ exhibit an exponential relationship. 
When the token length is less than $256$, the minimum number of codebooks $M$ increases rapidly, 
approaching the sample size of ImageNet (1281167 images). 
We fitted this curve and obtained the approximate relationship:
\begin{align}
    M \approx 10^{-3.6 \log_{10}{B} + 12.82}.
\end{align}
Based on this, we conclude that directly using the quantizer to quantize the latent vector of a trained VAE model has obvious performance limitations. 
Only by introducing additional nonlinear modules can the blue curve in \cref{fig:appendix_fit_curve} be shifted downward to achieve higher compression rates, 
which is a goal we hope to further pursue at the conclusion of this work.

\subsection{Detailed Results during Training} \label{sec:appendix_training}

\begin{figure}[H]
    \centering
    \begin{subfigure}[b]{0.48\textwidth}
        \includegraphics[width=\textwidth]{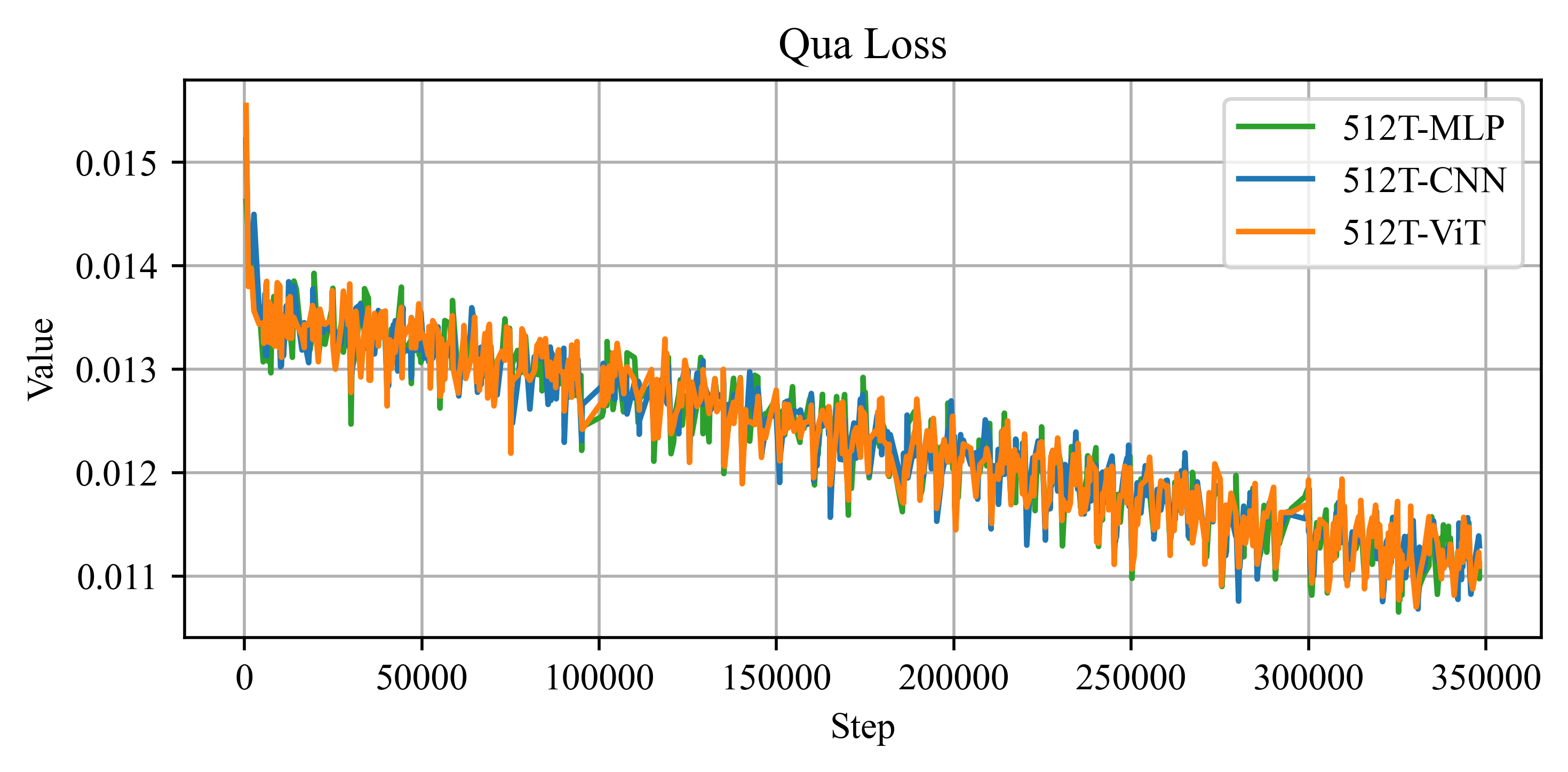}
        \caption{Quantizer loss with 512 tokens across different rectifier types.}
        \label{fig:training_details_a}
    \end{subfigure}
    \hfill
    \begin{subfigure}[b]{0.48\textwidth}
        \includegraphics[width=\textwidth]{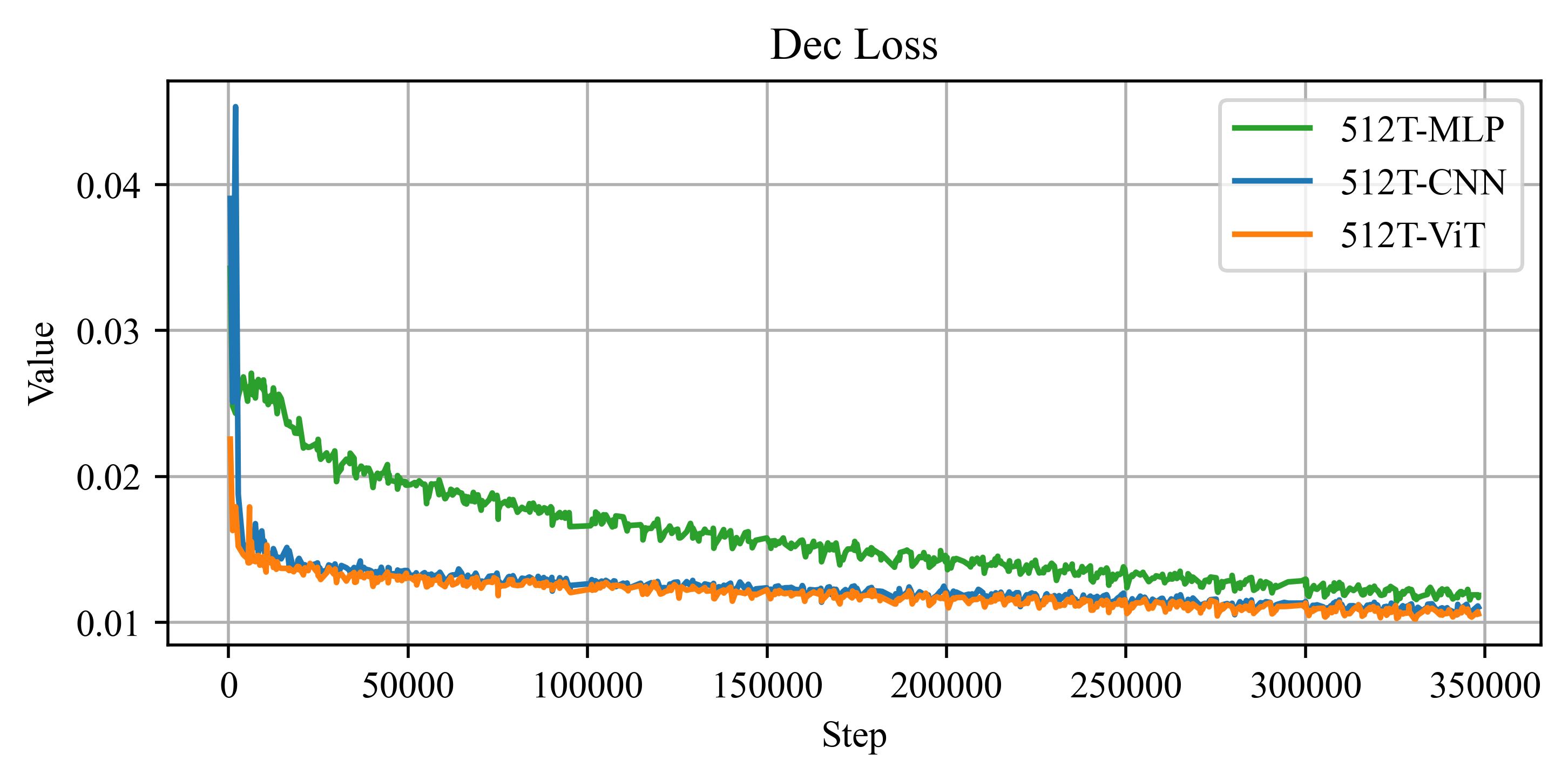}
        \caption{Decoder loss with 512 tokens across different rectifier types.}
        \label{fig:training_details_b}
    \end{subfigure}

    \vspace{1em}

    \begin{subfigure}[b]{0.48\textwidth}
        \includegraphics[width=\textwidth]{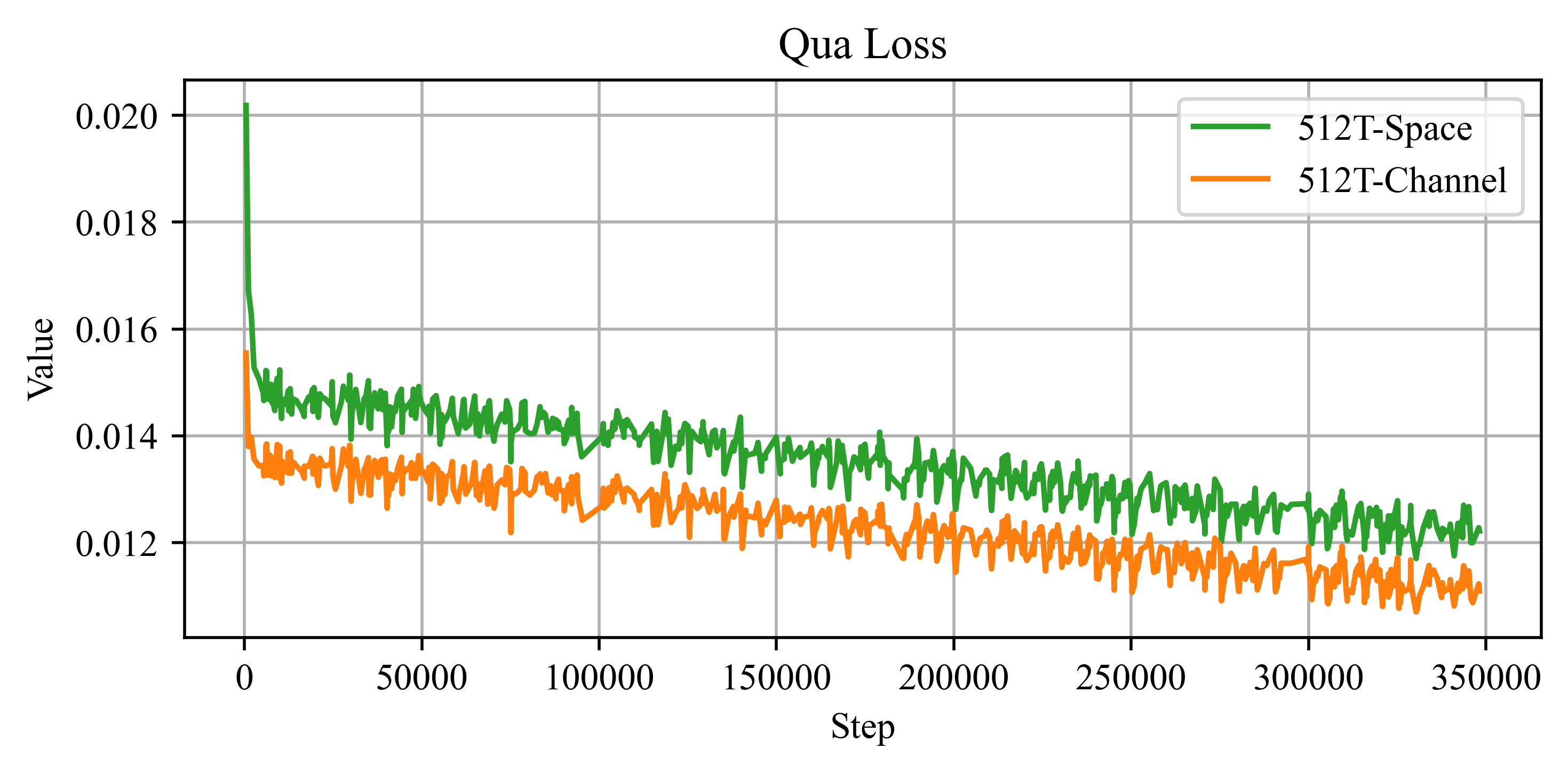}
        \caption{Quantizer loss with 512 tokens across different split types.}
        \label{fig:training_details_c}
    \end{subfigure}
    \hfill
    \begin{subfigure}[b]{0.48\textwidth}
        \includegraphics[width=\textwidth]{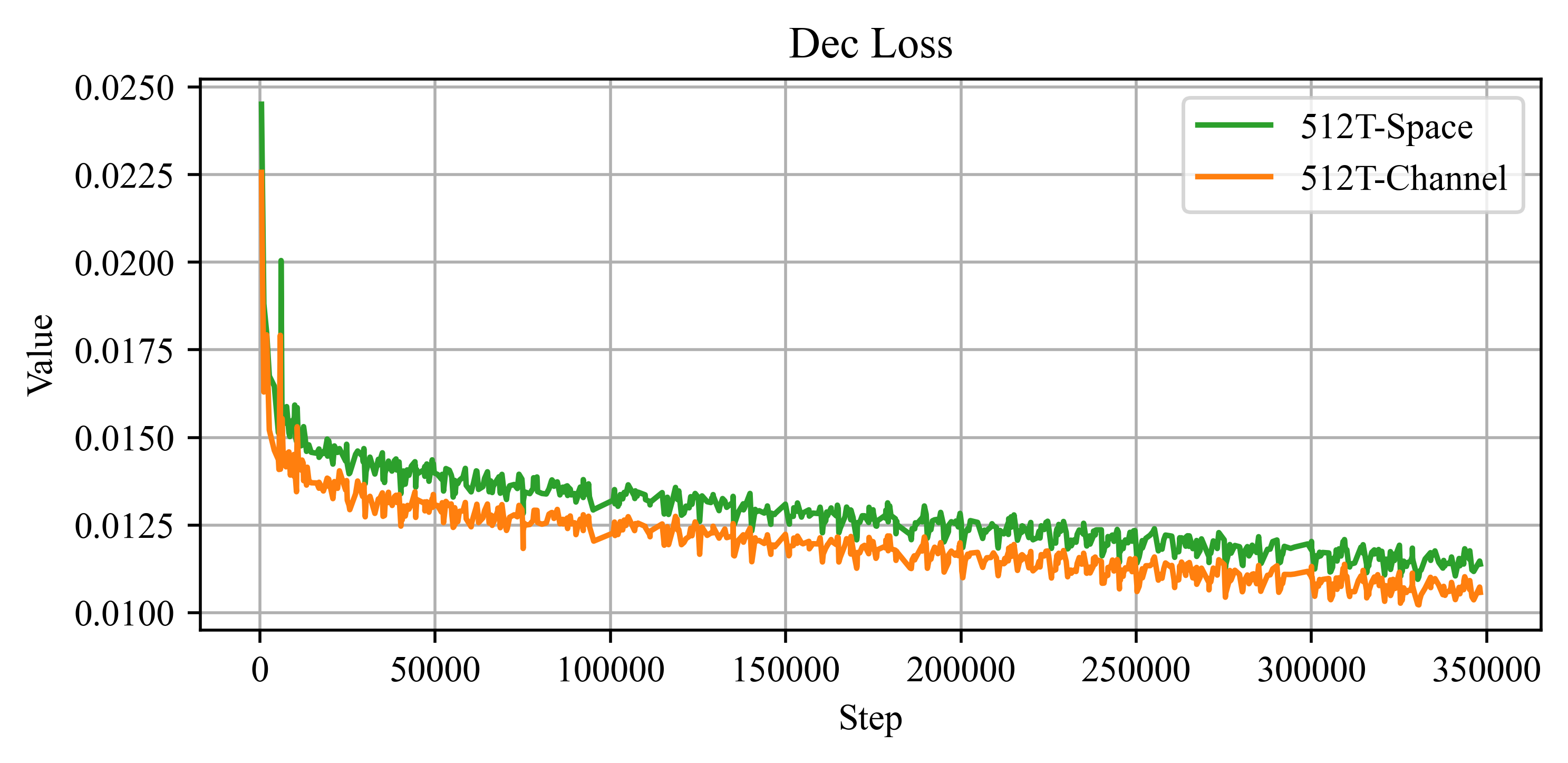}
        \caption{Decoder loss with 512 tokens across different split types.}
        \label{fig:training_details_d}
    \end{subfigure}

    \vspace{1em}

    \begin{subfigure}[b]{0.48\textwidth}
        \includegraphics[width=\textwidth]{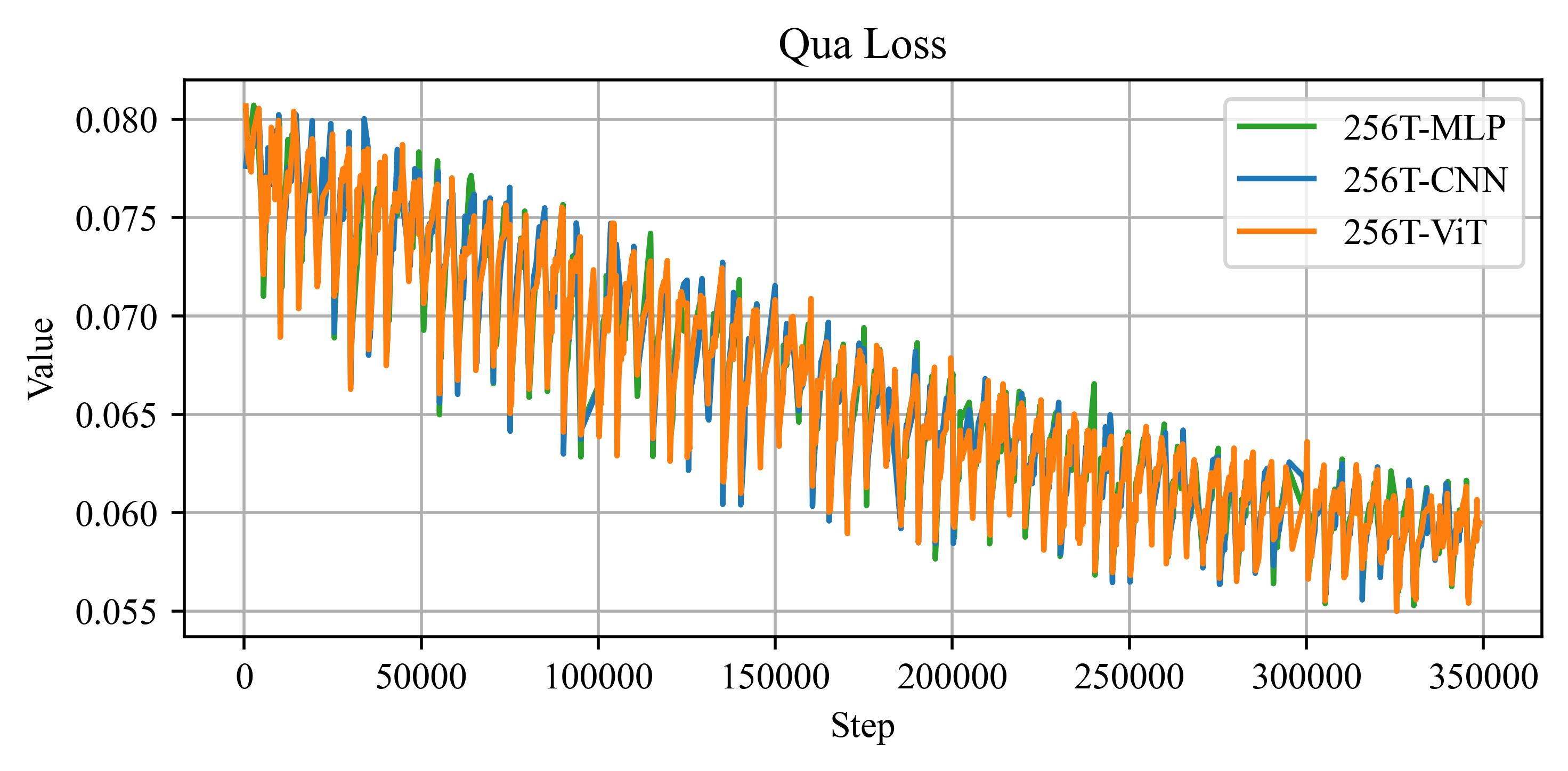}
        \caption{Quantizer loss with 256 tokens across different rectifier types.}
        \label{fig:training_details_e}
    \end{subfigure}
    \hfill
    \begin{subfigure}[b]{0.48\textwidth}
        \includegraphics[width=\textwidth]{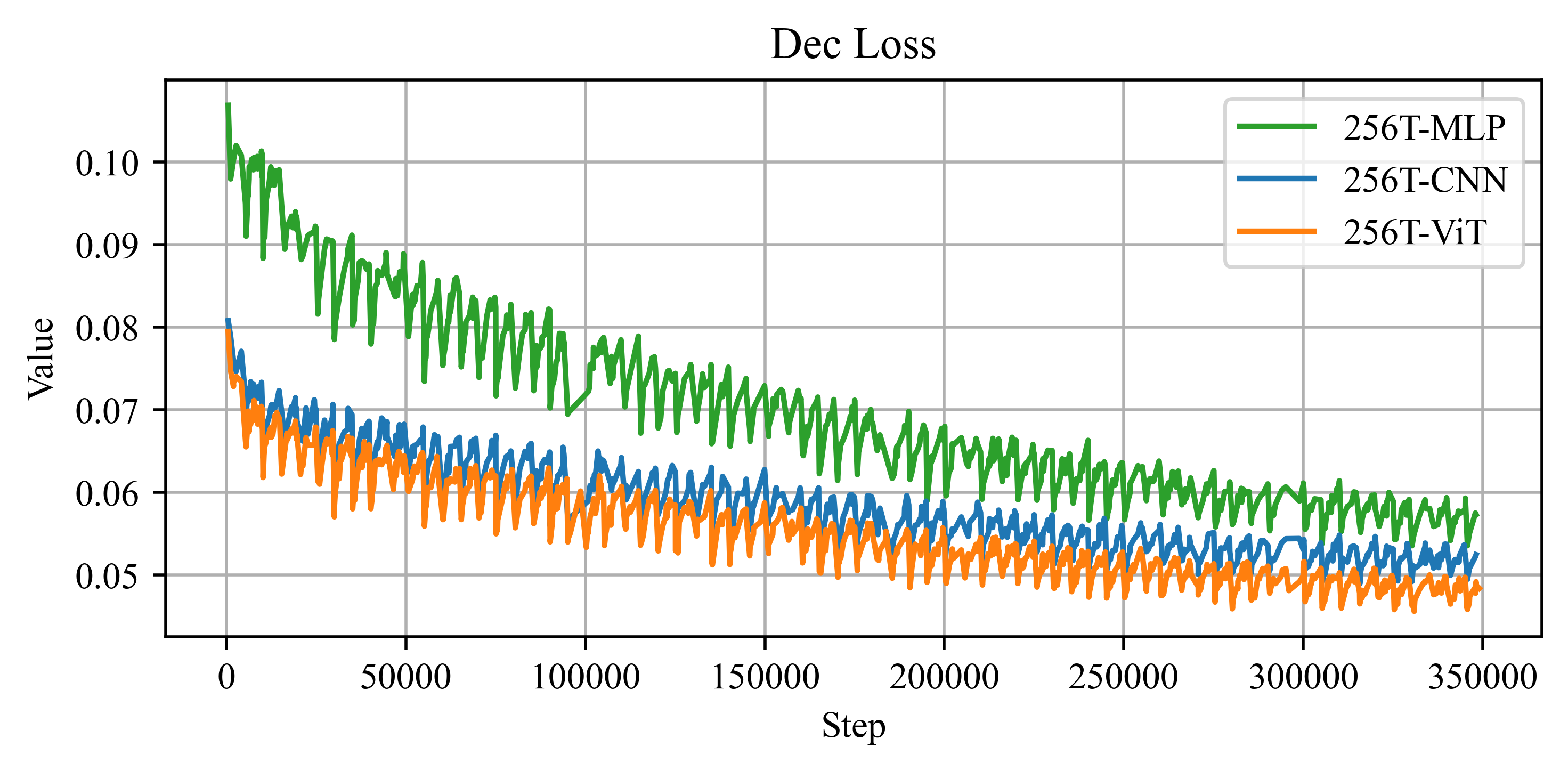}
        \caption{Decoder loss with 256 tokens across different rectifier types.}
        \label{fig:training_details_f}
    \end{subfigure}
    
    \vspace{1em}

    \begin{subfigure}[b]{0.48\textwidth}
        \includegraphics[width=\textwidth]{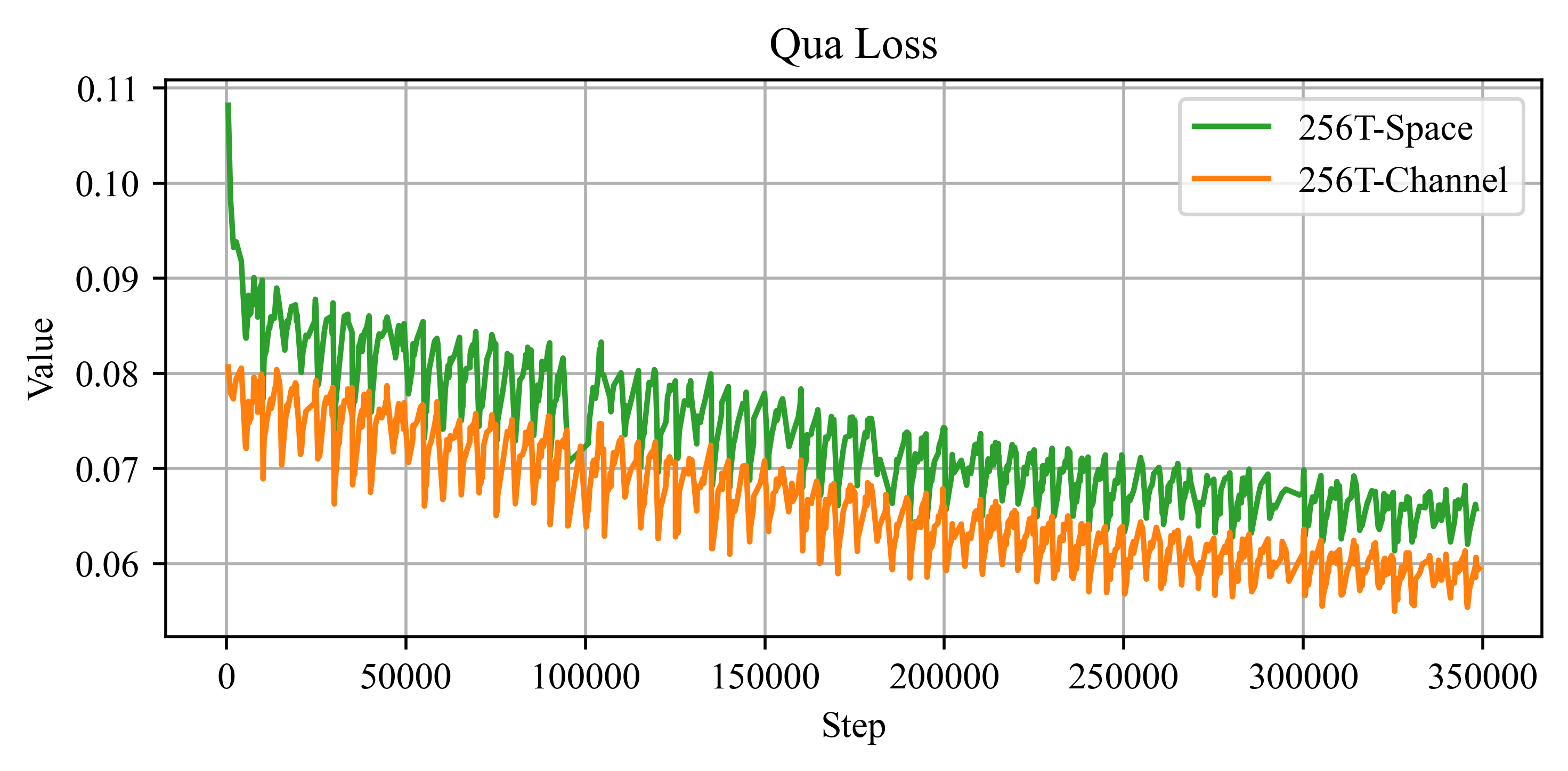}
        \caption{Quantizer loss with 256 tokens across different split types.}
        \label{fig:training_details_g}
    \end{subfigure}
    \hfill
    \begin{subfigure}[b]{0.48\textwidth}
        \includegraphics[width=\textwidth]{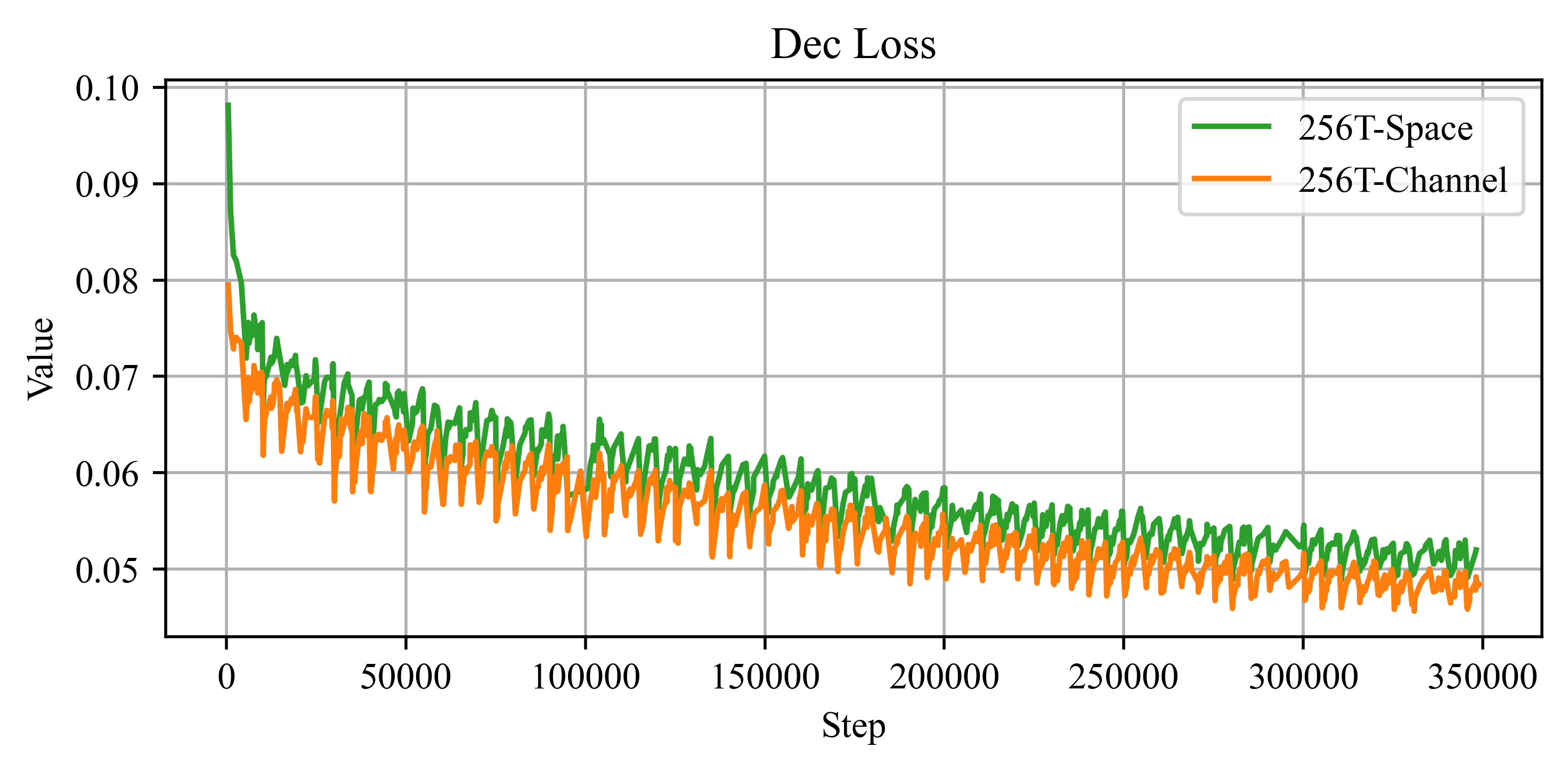}
        \caption{Decoder loss with 256 tokens across different split types.}
        \label{fig:training_details_h}
    \end{subfigure}
    
    \caption{Details training statistics.}
    \label{fig:training_details}
\end{figure}

To further demonstrate the effectiveness of the proposed ReVQ model, 
we present the overall loss curves recorded during training. 
As shown in \cref{fig:training_details_a,fig:training_details_b} 
(where the reconstruction error of quantizer features is abbreviated as ``Qua Loss'' and the reconstruction error of rectifier features is termed ``Dec Loss'' in the figures), 
the quantizer loss remains nearly unchanged, which is expected given that only the rectifier structure is varied while all other components are held constant. 
In contrast, the decoder loss is significantly affected by the rectifier type, with the ViT counterpart achieving the lowest loss.  
\cref{fig:training_details_c,fig:training_details_d} further illustrate that the splitting strategy has a substantial impact on both the quantizer and rectifier losses.
Specifically, the channel multi-group approach leads to consistently lower losses, indicating better overall model performance. \cref{fig:training_details_e,fig:training_details_f,fig:training_details_g,fig:training_details_h} provide additional validation for the above observations. 
Moreover, they reveal that using 256 tokens results in higher training loss compared to the 512-token configuration, 
suggesting that models with fewer tokens are more challenging to train. 
This observation implies that, for a given VAE architecture, the achievable compression ratio has an inherent upper bound.

\subsection{More Visualizations on ImageNet} \label{sec:appendix_imagenet}
\begin{figure}[tbp]
    \centering
    \includegraphics[width=0.83\linewidth]{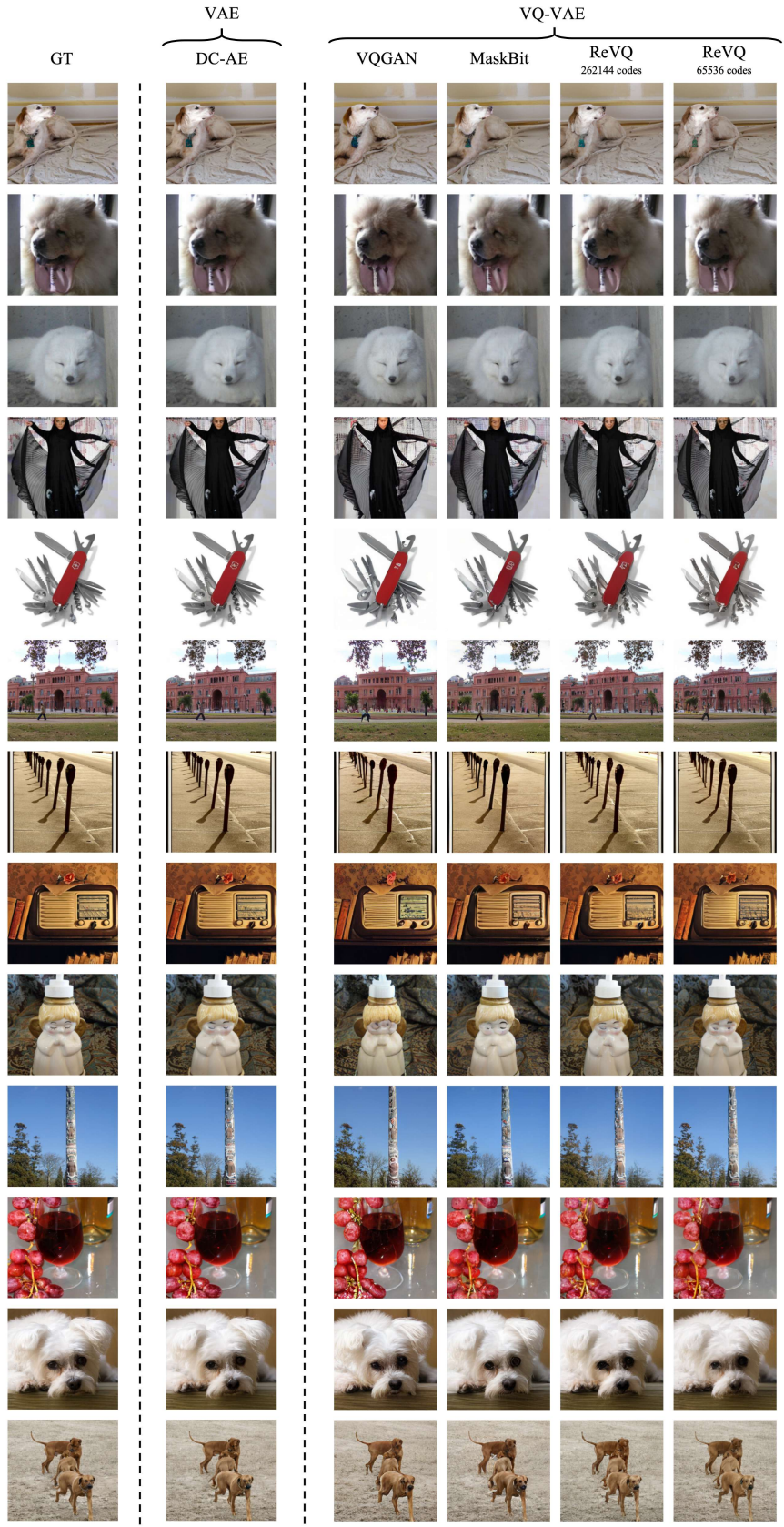}
    \caption{Additional reconstructed results on ImageNet dataset.}
    \label{fig:appendix_recon}
    \vspace{-4mm}
\end{figure}

In this section, we present additional reconstruction results, as shown in \cref{fig:appendix_recon}. 
These results further demonstrate the superior performance of the proposed ReVQ model on ImageNet dataset.

